\documentclass[11pt, a4paper, onecolumn, copyright, gr]{google}

\usepackage{amsmath,amsfonts,bm}

\def\eqref#1{equation~\ref{#1}}
\def\1{\bm{1}}

\DeclareMathAlphabet{\mathsfit}{\encodingdefault}{\sfdefault}{m}{sl}
\SetMathAlphabet{\mathsfit}{bold}{\encodingdefault}{\sfdefault}{bx}{n}

\usepackage[authoryear, sort&compress, round]{natbib}

\usepackage{hyperref}
\usepackage{url}

\usepackage{titletoc}
\usepackage{xspace}
\usepackage{tabularx, colortbl,booktabs}
\usepackage{enumitem}
\usepackage{listings}
\usepackage{pifont}
\usepackage{amssymb}
\usepackage{amsmath}
\uselogo{} 

\title{RECAST: Learning to Compute the Right Context through Adaptive Evidence Routing}

\reportnumber{} % Leave blank if n/a

\renewcommand{\today}{2026-10-07}

\author[1]{Yilun Hao$^{\dagger}$}
\author[2]{Krishna Sayana$^{*}$}
\author[2]{Isabella Ye$^{*}$}
\author[2]{James S Ren}
\author[2]{Sukhdeep Sodhi}
\author[2]{Craig Boutilier}
\author[1]{Chuchu Fan}

\affil[1]{MIT}
\affil[2]{\thepa{}{}}

\newcommand{\ours}{RECAST\xspace}

\usepackage[most]{tcolorbox}
\tcbuselibrary{listings,breakable}

\newtcblisting{benchmarkexample}[1]{
    enhanced,
    breakable,
    listing only,
    colback=gray!4,
    colframe=gray!45,
    colbacktitle=gray!15,
    coltitle=black,
    title=\textbf{#1},
    fonttitle=\small\bfseries,
    boxrule=0.5pt,
    arc=1mm,
    left=4pt,
    right=4pt,
    top=3pt,
    bottom=3pt,
    before skip=6pt,
    after skip=6pt,
    listing options={
        basicstyle=\ttfamily\footnotesize,
        breaklines=true,
        columns=fullflexible,
        keepspaces=true,
        showstringspaces=false,
        escapeinside={(*@}{@*)}
    }
}
\begin{document}

\begin{abstract}
Large language models are increasingly applied to tasks grounded in long,
heterogeneous information sources. Conventional Retrieval-Augmented Generation (RAG) relies on fixed
similarity-based retrieval, while agentic variants adapt queries and tool use
but remain largely retrieval-centric. However, in many tasks, the evidence required for a solution is not explicitly present in any single source item. Instead, it must be derived through filtering, aggregation, or computation across multiple source items. In this work, we introduce \ours
(\textbf{R}outing \textbf{E}vidence through \textbf{C}omputation,
\textbf{A}ccess, and \textbf{S}ynthesized \textbf{T}ools), a learned framework
that formulates evidence construction as a sequential decision process over
heterogeneous retrieval and computation operations, allowing evidence to be
actively derived rather than merely retrieved. A lightweight \texttt{RouterLM} iteratively selects and formulates primitive operations or specifies customized operations for a frozen \texttt{CompilerLM} to translate into executable code. Once it judges the evidence sufficient, \texttt{RouterLM} passes the accepted evidence to a frozen \texttt{AnswerLM} to produce the final solution. We train \texttt{RouterLM} with supervised fine-tuning (SFT) followed by group relative policy optimization (GRPO). Across six heterogeneous benchmark families, \ours achieves a mean success
rate of $75.6\%$, outperforming the strongest large-model baseline by
$15.9\%$. Moreover, training enables the Qwen3.5-9B \texttt{RouterLM} to
outperform a training-free Gemini~3.5 Flash \texttt{RouterLM} by $5.0\%$.
On three held-out benchmarks, \ours improves over the strongest baseline by
$15.0\%$ on average, demonstrating strong zero-shot generalization across
tasks and heterogeneous source representations.
\end{abstract}
\maketitle
\renewcommand{\thefootnote}{\fnsymbol{footnote}}
\footnotetext[1]{Equal contributions.}
\footnotetext[2]{Work done during internship at Google.}
\renewcommand{\thefootnote}{\arabic{footnote}} % (Optional) resets body footnotes back to numbers 1, 2...

\section{Introduction}
\label{sec:introduction}

Large language models (LLMs) are increasingly applied to diverse tasks, such
as question answering, data analysis, and personalized assistance, many of
which require reasoning over long contexts drawn from heterogeneous external
sources such as document collections, databases, tables, semi-structured
records, and user histories~\citep{nakano2021webgpt,jiang2023structgpt,
zhang2023data,li2023sheetcopilot,zhong2024memorybank,packer2023memgpt}.
Although modern LLMs can process increasingly long inputs, supplying the
complete source incurs substantial computational and token costs while still
offering no guarantee that the model will identify and correctly use the
information required for a particular task~\citep{jiang2024longllmlingua,
liu2024lost}. Retrieval-augmented generation (RAG) offers a practical solution
by retrieving a small amount of relevant content and conditioning generation
on the retrieved evidence~\citep{guu2020retrieval,lewis2020retrieval,
izacard2021leveraging}. Conventional RAG typically performs a single retrieval
step before generation, whereas iterative and agentic variants allow LLMs to
reformulate queries, invoke retrieval operations over multiple rounds, and
assess whether additional evidence is
needed~\citep{trivedi2023interleaving,hui2026interact}.

These retrieval-based formulations generally assume that useful evidence can be located directly within the source. However, LLM applications increasingly rely on heterogeneous sources, from which the required evidence may need to be derived rather than simply retrieved. For example, suppose the source provides monthly revenue and operating income, while the task asks which three-month period had the largest increase in operating margin. Answering the question requires computing the margins and comparing their changes across periods. More broadly, the required operations vary across tasks, from lexical or semantic retrieval to relational filtering and aggregation, and some tasks require customized computation that must be synthesized for the task at hand.

Context construction must
therefore determine not only \emph{where} relevant information appears, but
also \emph{how} it should be transformed into useful evidence. Although
iterative and agentic methods make retrieval more adaptive, they primarily
change how the system searches for source content and offer limited support for
constructing such derived evidence. We therefore consider a broader context-construction problem in which retrieval and computation serve as complementary evidence operations, adaptively selected and combined for each task instance.
\begin{figure*}[ht]
  \centering
  %\vspace{-5pt}
  \includegraphics[width=\linewidth]{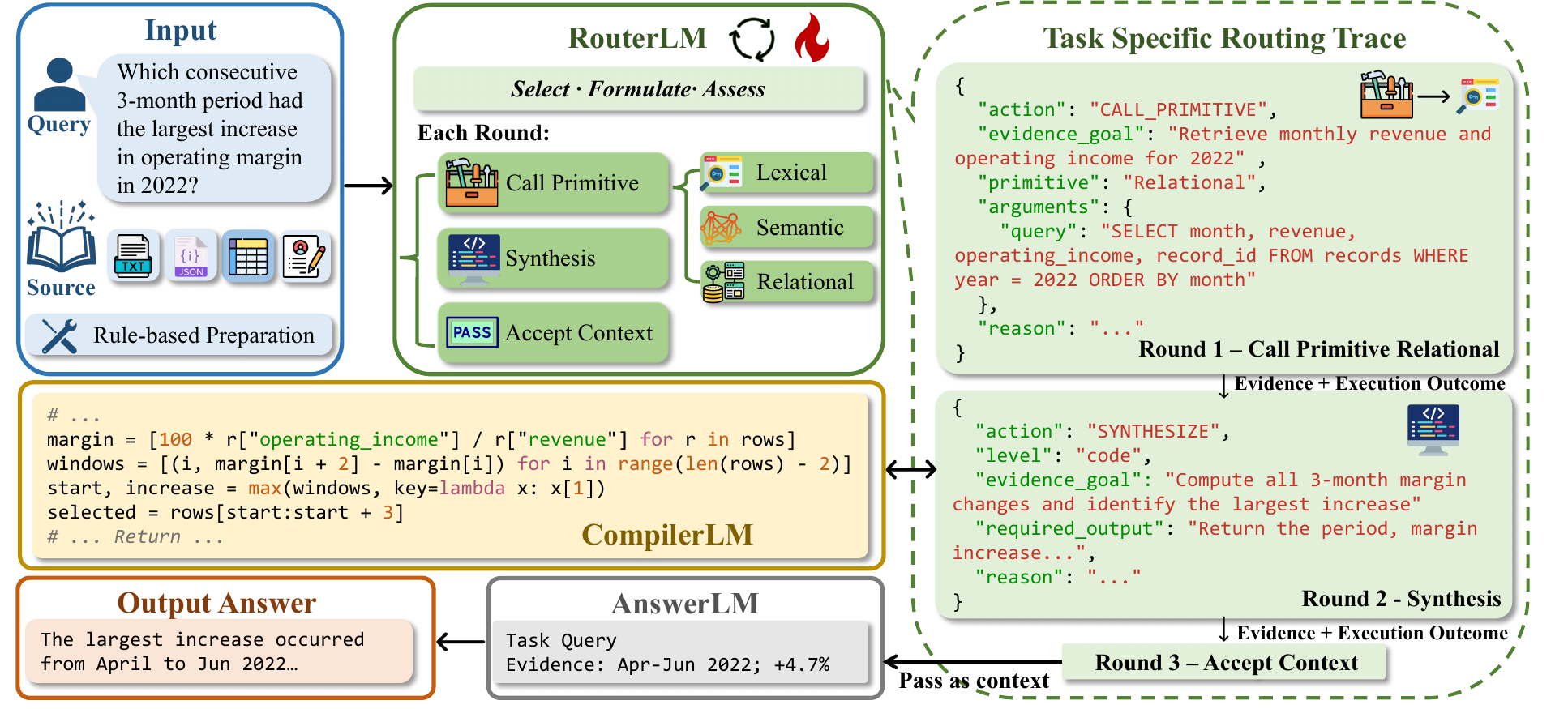} 
  %\vspace{-18pt}
\caption{\textbf{Overview of \ours.}
At each round, \texttt{RouterLM} uses the task, compact source profile, and
evidence and feedback from earlier rounds to either select and formulate a
primitive or synthesized operation or pass the evidence to
\texttt{AnswerLM} when it considers the evidence sufficient.}
  \label{fig:recast-overview}
  %\vspace{-15pt}
\end{figure*}

In this work, we introduce \ours
(\textbf{R}outing \textbf{E}vidence through \textbf{C}omputation,
\textbf{A}ccess, and \textbf{S}ynthesized \textbf{T}ools), illustrated in
Figure~\ref{fig:recast-overview}. \ours formulates evidence construction as a
learned sequential decision process over heterogeneous retrieval and
computation operations. At each round, a lightweight \texttt{RouterLM}
determines what evidence is needed and formulates a task-specific operation for
obtaining it. It may invoke a parameterized lexical, semantic, or relational
primitive, or request synthesis of a customized operation that a frozen
\texttt{CompilerLM} translates into executable code. The resulting evidence
and execution outcome inform subsequent decisions, allowing
\texttt{RouterLM} to refine the constructed evidence over multiple rounds.
Once the evidence is sufficient, \texttt{RouterLM} accepts it and passes it as
context to a frozen \texttt{AnswerLM} for final solution generation. We train the \texttt{RouterLM} using supervised fine-tuning (SFT) followed by group relative
policy optimization (GRPO). 

We evaluate \ours on DataBench, FinQA, HiTab, HotpotQA, LaMP, and
MultiHiertt~\citep{grijalba2024question,chen2021finqa,cheng2022hitab,
yang2018hotpotqa,salemi2024lamp,zhao2022multihiertt}. These six benchmarks span heterogeneous source representations, including dataframes,
financial reports, hierarchical tables, Wikipedia passage collections, and
user profiles. Across them, \ours achieves an average success rate of $75.6\%$,
outperforming the strongest large-model baseline at $59.7\%$. The trained Qwen3.5-9B
\texttt{RouterLM} outperforms a large training-free Gemini~3.5 Flash
router, which achieves $70.6\%$. On three held-out benchmarks,
2WikiMultiHopQA, TAT-QA, and WikiTableQuestions~\citep{ho2020constructing,
zhu2021tat,pasupat2015compositional}, \ours obtains $79.3\%$, compared with
$64.3\%$ for the strongest baseline, demonstrating strong zero-shot
generalization across tasks and heterogeneous source representations.

In summary, our key contributions are:
\begin{itemize}
    \item We broaden evidence construction beyond retrieval by treating retrieval and computation as complementary means of obtaining task-relevant evidence from heterogeneous sources.

    \item We propose \ours, a multi-round framework in which a lightweight \texttt{RouterLM} formulates primitive or synthesized evidence operations, assesses their outcomes, and determines when evidence is sufficient. We further improve its evidence-construction policy through SFT and GRPO.

    \item We evaluate \ours across 6 in-domain and 3 held-out benchmark families. \ours achieves $75.6\%$ in-domain and $79.3\%$ on held-out benchmarks, outperforming the strongest baselines by $15.9\%$ and $15.0\%$, respectively. It also surpasses a large training-free \texttt{RouterLM} and transfers effectively across alternative \texttt{CompilerLM}s.
\end{itemize}

\section{Related Works}
\label{sec:related_work}

\paragraph{Retrieval-Augmented and Agentic Context Construction.}
Retrieval-augmented generation (RAG) grounds language-model outputs in external information by retrieving relevant source content and conditioning generation on it~\citep{guu2020retrieval,karpukhin2020dense,lewis2020retrieval,
izacard2021leveraging}. Subsequent work develops iterative and hierarchical retrieval strategies for multi-hop and long-context reasoning, such as evidence condensation across retrieval hops, reasoning-guided query formulation, and retrieval over hierarchical document representations~\citep{khattab2021baleen,trivedi2023interleaving,sarthi2024raptor}.
More adaptive systems decide when and how to retrieve, assess or correct retrieved results, and use intermediate reasoning to guide subsequent retrieval actions~\citep{jiang2023active,asai2024self,jeong2024adaptive,
yan2024corrective,yao2022react,hui2026interact}. While these methods move beyond fixed one-pass retrieval, they still construct evidence primarily by selecting or refining information already present in the source. \ours instead
extends adaptive context construction to relational computation and synthesized transformations that derive new evidence from source content.

\paragraph{Tool-Augmented and Programmatic Reasoning.}
A complementary line of work augments language models with structured
operations, tools, and executable programs. Methods for structured
sources use specialized table models, symbolic programs, or iterative
operations over tables and records
\citep{herzig2020tapas,liu2021tapex,jiang2023structgpt,
cheng2022binding,wang2024chain}. General tool-use frameworks allow language
models to select and invoke predefined tools
\citep{schick2023toolformer,paranjape2023art,qin2024toolllm}, while
program-aided and tool-synthesis approaches generate executable programs or
new tools for the task
\citep{gao2023pal,chen2022program,qian2023creator,cai2024large,
gemmell2023toolwriter}. These methods broaden the computational capabilities of language models, but generally either generate programs that directly produce task answers or rely on operations designed for a particular source type. \ours instead learns to construct task-relevant evidence across heterogeneous sources by iteratively selecting, formulating, and evaluating primitive or synthesized operations before final solution generation.

\section{\ours}
\label{sec:method}

% The goal of \ours is to construct task-relevant evidence from long and heterogeneous sources.
% A trainable \texttt{RouterLM} iteratively determines how evidence should be
% accessed or computed, while frozen auxiliary models support code generation
% and final answer generation. We next formulate the evidence-construction
% problem, describe the \ours pipeline, and present its two-stage training
% procedure.

\subsection{Problem Formulation and Challenges}
\label{sec:problem}
Given a natural-language task $q$ and an associated external source collection
$\mathcal{X}=\{x_i\}_{i=1}^{N}$, our goal is to construct query-specific
evidence that enables an LLM to produce the solution $y$. We write this
process as
\begin{equation}
    E = G_{\theta}(q,\mathcal{X}), \qquad
    y = F_{\mathrm{ans}}(q,E),
\end{equation}
where $G_{\theta}$ denotes the evidence-construction process parameterized by
the learned routing policy and $F_{\mathrm{ans}}$ denotes the frozen answer
model. A source item $x_i$ may take diverse forms, such as a text passage,
structured table, JSON record, hierarchical table, or user-profile entry.
Rather than applying $F_{\mathrm{ans}}$ directly to the complete source,
\ours constructs compact evidence $E$ by adaptively retrieving, transforming,
or computing information from $\mathcal{X}$. This evidence may include both
directly retrieved source content and information derived through filtering,
aggregation, arithmetic, linking, or other transformations across multiple
source items. This setting presents two main challenges:

% Given a natural-language task $q$ and its associated source context
% $\mathcal{X}=\{x_i\}_{i=1}^{N}$, our goal is to produce a solution $y$. A
% source item $x_i$ may take diverse forms, such as a text passage, structured
% table, JSON record, hierarchical table, or user-profile entry. Rather than
% providing the entire source context directly for solution generation, \ours constructs compact,
% task-relevant evidence $E$ by adaptively retrieving, transforming, or computing
% information from $\mathcal{X}$. The constructed evidence may consist not only of directly retrieved source content, but also of information derived through filtering, aggregation, arithmetic, linking, or other transformations across multiple source items.
% This setting presents two main challenges.

$\bullet$ \textbf{Operation Diversity and Formulation.} Different tasks require fundamentally different evidence operations. Finding exact names or identifiers may require lexical matching, locating paraphrased concepts may require semantic retrieval, and combining or calculating over multiple records may require relational or algorithmic computation. When the required transformation falls outside the capabilities of reusable
retrieval and computation operations, a task-specific program may be needed. Selecting an operation alone is
insufficient. The system must also formulate the task-specific request by
specifying the query or computation, the records to use, and the
evidence to return.

$\bullet$ \textbf{Sequential Decisions and Evidence Assessment.} Evidence
construction may require multiple rounds. An initial operation may identify
the subset of the source needed by a later operation, return insufficient
evidence, or fail at runtime. The system must assess each outcome,
decide whether another operation is needed, and stop only when the retained
evidence supports final inference.

% We formulate evidence construction as a multi-round policy.  At round $t$,
% \texttt{RouterLM} observes the task, a deterministic source profile $m$, the
% retained evidence $E_t$, and execution feedback $F_t$.  It then samples an
% action
% \begin{equation}
%     a_t \sim \pi_{\theta}(\cdot \mid q,m,E_t,F_t),
% \end{equation}
% where $a_t$ either calls a primitive evidence operation, requests code
% synthesis, or accepts the current context.  Evidence-producing actions return
% new evidence $\Delta E_t$ and structured feedback, after which
% $E_{t+1}=E_t\cup\Delta E_t$.  Selecting the acceptance action terminates the
% routing loop and invokes \texttt{AnswerLM}.

\subsection{\ours Overview}
\label{sec:overview}

% To address these challenges, we propose \ours
% (Figure~\ref{fig:recast-overview}), a multi-round framework that adaptively
% constructs task-relevant evidence by selecting and formulating retrieval or
% computation operations and assessing their results. \ours uses a trainable
% \texttt{RouterLM} to control evidence construction, a frozen
% \texttt{CompilerLM} to implement evidence operations that require customized
% code, and a frozen \texttt{AnswerLM} to produce the final solution. The
% framework proceeds in four steps.

To address these challenges, we propose \ours
(Figure~\ref{fig:recast-overview}), a multi-round implementation of the
evidence-construction mapping $G_{\theta}$ introduced above.
\texttt{RouterLM}, parameterized by $\theta$, implements the policy that
selects, formulates, and assesses evidence operations. A frozen
\texttt{CompilerLM}, denoted $F_{\mathrm{comp}}$, translates customized
operation specifications into executable code, while the frozen
\texttt{AnswerLM} implements $F_{\mathrm{ans}}$ and maps the constructed
evidence to the final solution. Only \texttt{RouterLM} is trained. The
framework proceeds in four steps.

\noindent\textbf{Step 1: Source preparation.}
Because source information may be represented in different formats, a
deterministic, rule-based preprocessor converts the source context
$\mathcal{X}$ into a uniformly formatted list of records and constructs a
compact source profile $m$. The profile summarizes the source scale, structure, formats, and observed fields, providing \texttt{RouterLM} with a high-level view of the source and allowing it to operate across heterogeneous sources without receiving the complete source in its prompt. This rule-based preparation uses no language model and is independent of the task. As an initialization step, both the evidence set and interaction history are empty:
$E_0=\emptyset$ and $H_0=\emptyset$.

\noindent\textbf{Step 2: Evidence-operation formulation.}
At round $t$, \texttt{RouterLM} observes the task $q$, source profile $m$,
retained evidence $E_t$, and interaction history $H_t$. It selects an action
according to
\begin{equation}
    a_t \sim \pi_{\theta}(\cdot \mid q,m,E_t,H_t).
\end{equation}
The action either invokes a primitive operation from $\{\texttt{Lexical},\texttt{Semantic},\texttt{Relational}\}$,
requests a synthesized evidence operation, or accepts the current evidence. Section~\ref{sec:operations} describes these operations in detail. This step does not produce the final task solution. Instead,
\texttt{RouterLM} must assess what evidence is still needed, select an
appropriate operation, and formulate its task-specific request, including the
arguments or output requirements that specify how the evidence should be
obtained and what should be returned.

\noindent\textbf{Step 3: Operation execution and feedback.}
For a primitive action, the selected operation is executed using task-specific arguments formulated by \texttt{RouterLM}. A
synthesis request is translated by \texttt{CompilerLM} into a Python program and then executed over the source. Either path returns an operation outcome
$o_t$ describing execution status, result counts, and any errors.
A successful operation may return new evidence $\Delta E_t$. The
evidence and interaction states are updated as
% \noindent\textbf{Step 3: Operation execution and feedback.}
% For a primitive action, the selected operation is executed using the
% task-specific arguments formulated by \texttt{RouterLM}. A
% synthesis request is translated by \texttt{CompilerLM} into a Python
% program, which is then executed over the source. Either path returns an operation outcome $o_t$ describing its execution status, result counts, and any errors. 
% % A successful operation additionally returns new evidence $\Delta E_t$. The evidence and interaction states are updated as\begin{equation}E_{t+1}=E_t\cup\Delta E_t,\qquad H_{t+1}=H_t\cup{(a_t,o_t)}.\end{equation}
% A successful operation may additionally return new evidence $\Delta E_t$. The
% evidence and interaction states are updated as
\begin{equation}
    E_{t+1}=E_t\cup\Delta E_t,
    \qquad
    H_{t+1}=H_t\mathbin{\|}[(a_t,o_t)],
\end{equation}

where $\|$ denotes appending the latest action and outcome to the interaction
history.

\noindent\textbf{Step 4: Evidence assessment and solution generation.}
Steps 2--3 are repeated until \texttt{RouterLM} considers the accumulated
evidence sufficient. \texttt{RouterLM} then accepts $E_t$ as the final evidence and terminates the
evidence-construction loop.
Finally, \texttt{AnswerLM} is invoked once with the original task $q$ and the
accepted evidence $E_t$ to produce the solution $y$.

\begin{figure*}[ht]
  \centering
  %\vspace{-5pt}
  \includegraphics[width=\linewidth]{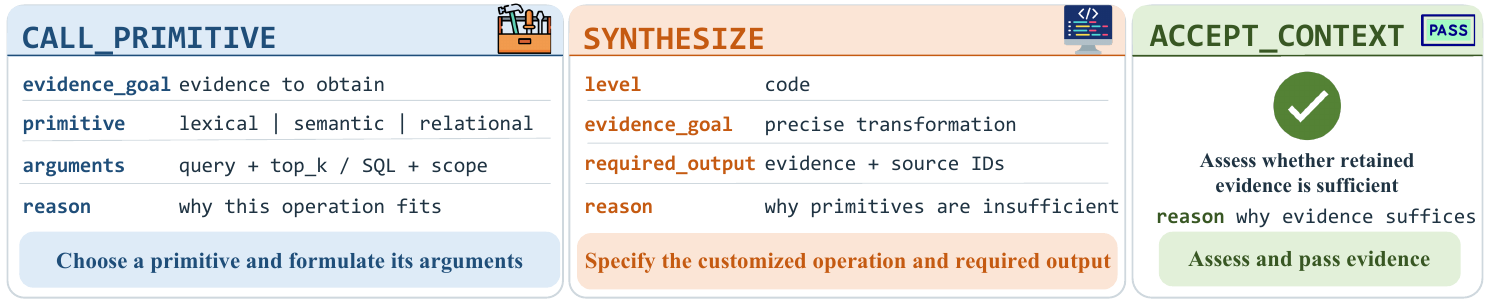} 
  %\vspace{-18pt}
  \caption{\textbf{\texttt{RouterLM} action space and output structure.}
Together, the outputs represent action selection, task-specific formulation,
and justification.}
  \label{fig:recast-actions}
  %\vspace{-10pt}
\end{figure*}

Figure~\ref{fig:recast-actions} summarizes the structured action space of
\texttt{RouterLM}. At each round, \texttt{RouterLM} selects one of three
actions and formulates the task-specific information needed to carry it out.
\texttt{CALL\_PRIMITIVE} and \texttt{SYNTHESIZE} construct additional
evidence, whereas \texttt{ACCEPT\_CONTEXT} terminates evidence construction
and proceeds to final solution generation.

\subsection{Evidence Operations}
\label{sec:operations}
\subsubsection{CALL\_PRIMITIVE}
\paragraph{Operation selection and formulation.}
\texttt{RouterLM} selects an operation from
$\{\texttt{Lexical}$, $\texttt{Semantic}$, $\texttt{Relational}\}$.
\texttt{Lexical} ranks source records using BM25 term
matching~\citep{robertson2009probabilistic}, while \texttt{Semantic} ranks
them by cosine similarity using BGE-M3
embeddings~\citep{chen2024bge}. \texttt{Relational} executes a read-only
Structured Query Language (SQL) query over the normalized records, supporting
filtering, joins, grouping, aggregation, sorting, and arithmetic.
\texttt{RouterLM} must also formulate the selected operation for the current
evidence goal. Lexical and semantic calls specify a search query and the
number of records to return, whereas relational calls specify a SQL query. A
call may additionally be restricted to source records identified in earlier
rounds.

\paragraph{Execution and output.}
The formulated operation is executed directly over the source records and
returns retrieved content or a computed result together with its source
identifiers. The executor also reports whether the operation succeeded, how
many records matched and were returned, and any execution errors. Successful
execution indicates that the requested operation ran correctly, but does not
imply that the resulting evidence is sufficient for the task.

\subsubsection{SYNTHESIZE}

\paragraph{Operation specification.}
When the required transformation cannot be expressed through the
available primitives, \texttt{RouterLM} requests a synthesized operation. \texttt{RouterLM} specifies the
transformation to perform, the evidence and source identifiers to return, and
why the primitive operations are insufficient. Because \texttt{CompilerLM}
does not receive the original task, the specification must preserve the
task-specific constraints required to implement the operation correctly.

\paragraph{Compilation and execution.}
\texttt{CompilerLM} receives the synthesis specification, source profile, source samples, and prior operation outcomes, and translates the specification
into a Python program in one call. The program can process the complete
source, use previously obtained evidence, and invoke primitive
operations when useful. It returns either the requested evidence or an
execution error to \texttt{RouterLM}. 
In Figure~\ref{fig:recast-overview} example, \texttt{RouterLM} specifies the computation for identifying the three-month period with the largest margin increase, while \texttt{CompilerLM} translates that specification into code.
% Thus, \texttt{RouterLM} determines and formulates the required operation, while \texttt{CompilerLM} implements it.

\subsubsection{ACCEPT\_CONTEXT}

\paragraph{Evidence assessment.}
After each evidence-producing action, \texttt{RouterLM} receives the returned
evidence and structured feedback generated by the operation executor. The
feedback reports the execution status, result counts, and any errors. The
evidence is deduplicated and added to the current evidence set, while the
action and feedback are recorded in the interaction history.
\texttt{RouterLM} then assesses whether another operation is needed. If so, it
may reformulate the request or select a different operation in the next round.

\paragraph{Context acceptance.}
Once \texttt{RouterLM} determines that the constructed evidence is sufficient
for final inference, it selects \texttt{Accept\_Context}. This terminates
evidence construction and passes the constructed evidence to
\texttt{AnswerLM} for solution generation.

\subsection{\ours Training}
\label{sec:training}

In \ours, training updates only \texttt{RouterLM}. The primitive operations are deterministic, and the parameters of \texttt{CompilerLM} and \texttt{AnswerLM} remain fixed. Training
has two stages. We apply supervised fine-tuning (SFT) to
establish valid multi-round evidence-construction behavior, including
operation selection, task-specific formulation, and evidence assessment. We
apply Group Relative Policy Optimization (GRPO) to improve the routing
policy using downstream task outcomes. The complete dataset and training details of SFT and GRPO are provided in
Appendix~\ref{sec:appendix_ours_training}.

% \paragraph{Trajectory collection and SFT.}
% Using the pretrained \texttt{RouterLM}, we generate multiple candidate
% trajectories for each training question. Each trajectory executes the selected
% evidence operations and invokes \texttt{AnswerLM} with the constructed
% evidence. A frozen LLM judge then evaluates the resulting solution against the
% reference answer. For each question with at least one correct, executable, and
% structurally valid trajectory, we select one trajectory as the SFT expert,
% preferring more accurate and efficient candidates when multiple trajectories
% qualify. We apply capped upsampling by repeating examples from underrepresented
% dataset sources and empirical difficulty levels.
% The complete expert trajectory is used for SFT, supervising each
% \texttt{RouterLM} decision conditioned on the evidence and feedback available
% at that round.

\paragraph{Trajectory collection and SFT.}
Using the pretrained \texttt{RouterLM}, we generate multiple trajectories for
each training question. Each trajectory executes its evidence operations and
invokes \texttt{AnswerLM} with the constructed evidence. A frozen LLM judge
then evaluates the solution against the reference. For each question with at
least one correct, executable, and structurally valid trajectory, we select one
trajectory as the SFT expert, preferring more accurate and efficient
candidates. We apply capped upsampling by repeating examples from
underrepresented benchmark families and empirical difficulty levels. The
complete expert trajectory is used for SFT, supervising each \texttt{RouterLM}
decision conditioned on the evidence and feedback available at that round.

\paragraph{GRPO.}
We construct the GRPO data from questions that were not consistently solved
during SFT trajectory collection. Each selected question must have at least
one valid expert trajectory, indicating that it is solvable, and at least one
failed trajectory, indicating room for policy improvement. Starting from the
SFT checkpoint, we sample a new group of four complete trajectories for each
selected question. We then apply \emph{mixed-outcome filtering}, retaining only
groups containing at least one successful and one unsuccessful trajectory.
This allows GRPO to compare alternative evidence-construction behaviors for
the same question and reinforce those that produce better task outcomes.

For a trajectory $\tau$, we define the reward as
\begin{equation}
    R(\tau)=\lambda_C C(\tau)+\lambda_F F(\tau)+\lambda_V V(\tau),
\end{equation}
where $C(\tau)$ denotes final-answer correctness evaluated by the frozen LLM
judge, $F(\tau)$ is token-level F1 between the generated and reference answers,
and $V(\tau)$ indicates whether the trajectory follows the required action
structure. We set $\lambda_C=0.90$, $\lambda_F=0.08$, and
$\lambda_V=0.02$, prioritizing task correctness while retaining smaller
signals for partial answer quality and structural validity.

\section{Experimental Results}
\label{sec:experiments}

\subsection{Experimental Setup}
\label{sec:setup}

\paragraph{Benchmarks.}
We train and evaluate \ours on six benchmark families with heterogeneous
source representations. \texttt{DataBench} pairs questions with dataframes
from diverse domains~\citep{grijalba2024question}. \texttt{FinQA} contains
financial reports combining prose passages and tables~\citep{chen2021finqa}.
\texttt{HiTab} provides tables with hierarchical row and column
headers~\citep{cheng2022hitab}, while \texttt{MultiHiertt} associates questions
with multiple hierarchical tables and passages from financial
reports~\citep{zhao2022multihiertt}. \texttt{HotpotQA} presents each question
with Wikipedia passages~\citep{yang2018hotpotqa}. \texttt{LaMP} pairs an input
with a user profile of historical items for
personalization~\citep{salemi2024lamp}. We test on 100 unseen questions per
family.

% \paragraph{Benchmarks.}
% We train and evaluate on six benchmark families with heterogeneous sources.
% \texttt{DataBench} pairs questions with dataframes from diverse
% domains~\citep{grijalba2024question}. \texttt{FinQA} combines financial text
% and tables~\citep{chen2021finqa}. \texttt{HiTab} provides tables with
% hierarchical headers~\citep{cheng2022hitab}, while \texttt{MultiHiertt}
% contains financial reports with multiple hierarchical tables and
% passages~\citep{zhao2022multihiertt}. \texttt{HotpotQA} presents each question
% with Wikipedia passages~\citep{yang2018hotpotqa}. \texttt{LaMP} pairs an input
% with a user profile of historical items for
% personalization~\citep{salemi2024lamp}. We test on 100 unseen questions per
% family.

\paragraph{Held-out benchmarks.}
To evaluate zero-shot generalization, we use three benchmarks excluded from
SFT, GRPO, and checkpoint selection. \texttt{2WikiMultiHopQA} pairs questions
with collections of Wikipedia passages~\citep{ho2020constructing}.
\texttt{TAT-QA} contains financial reports combining prose passages and
tables~\citep{zhu2021tat}, while \texttt{WikiTableQuestions} pairs
natural-language questions with semi-structured Wikipedia
tables~\citep{pasupat2015compositional}. We evaluate on 100 examples per
benchmark. All nine benchmark descriptions and examples appear in
Appendices~\ref{sec:appendix_bench_desc} and~\ref{sec:appendix_bench_ex}.

% \paragraph{Held-out benchmarks.}
% To evaluate zero-shot generalization, we additionally use three benchmarks
% excluded from SFT, GRPO, and checkpoint selection.
% \texttt{2WikiMultiHopQA} pairs questions with collections of Wikipedia
% passages~\citep{ho2020constructing}. \texttt{TAT-QA} contains financial reports
% combining prose passages and tables~\citep{zhu2021tat}, while
% \texttt{WikiTableQuestions} pairs natural-language questions with
% semi-structured Wikipedia tables~\citep{pasupat2015compositional}. We evaluate on 100 examples from each held-out benchmark. Both in-domain and held-out benchmark descriptions and examples appear in Appendix~\ref{sec:appendix_bench_desc} and~\ref{sec:appendix_bench_ex}.

\paragraph{Baselines.}
We compare \ours against five baseline families spanning increasingly
expressive forms of context construction: \textbf{(1) Direct} provides only the
task to \texttt{AnswerLM}, without accessing the source;
\textbf{(2) Fixed Retrieval} uses the task as the query for a single BGE-M3
dense retrieval and returns the top three source chunks; \textbf{(3) Direct
Code} generates and executes a single Python program over the source without
iterative evidence construction; \textbf{(4) IRCoT}~\citep{trivedi2023interleaving} alternates sentence-level
reasoning with BGE-M3 retrieval; and
\textbf{(5) Interact-RAG}~\citep{hui2026interact} uses a
planner--reasoner--executor workflow to formulate retrieval actions and decide
when to stop. IRCoT and Interact-RAG are adapted to the per-instance sources and shared
model and retrieval components used in our evaluation; further implementation
details are provided in Appendix~\ref{sec:appendix-baselines}. For Direct Code, IRCoT, and Interact-RAG, we evaluate Qwen3.5-9B
and Gemini~3.5 Flash without task-specific fine-tuning as their respective
program-generation or retrieval-guidance LMs.

\paragraph{\ours variants.}
We evaluate \ours along five dimensions: \textbf{(1) \texttt{RouterLM} configurations} compare training-free
Qwen3.5-9B and Gemini~3.5 Flash routers with an SFT+GRPO-trained Qwen3.5-9B
router;
\textbf{(2) evidence-operation space} restricts training-free \ours to either
primitive operations or synthesized code while retaining full
multi-round routing; \textbf{(3) GRPO reward and data selection}
replaces the shaped reward with answer correctness alone or removes
mixed-outcome filtering; \textbf{(4) training stages} applies SFT and GRPO
independently; and \textbf{(5) compiler transfer} replaces
\texttt{CompilerLM} with alternative models. Details are in
Appendix~\ref{sec:appendix-ablations}.

% \paragraph{Implementation and Evaluation.}
% Semantic retrieval uses BGE-M3, lexical retrieval uses BM25, and relational
% operations execute read-only SQLite queries. All methods use Gemini~3.5 Flash
% as the shared \texttt{AnswerLM} and correctness judge, with an identical
% evaluation prompt. For \ours, Gemini~3.5 Flash also serves as
% \texttt{CompilerLM}. Qwen3.5-9B is evaluated with its thinking mode disabled.
% The training data contain 3,368 SFT trajectories and 608 GRPO training
% instances. We train \texttt{RouterLM} using Low-Rank Adaptation
% (LoRA)~\citep{hu2021lora}. We repeat each in-domain experiment three times
% and report mean task success rates, with standard deviations provided in
% Appendix~\ref{sec:appendix-main-std}. Additional implementation and training
% details are provided in Appendix~\ref{sec:appendix_ours}.
\paragraph{Implementation and Evaluation.}
% Semantic retrieval uses BGE-M3, lexical retrieval uses BM25, and relational
% operations execute read-only SQLite queries. For \ours, Gemini~3.5 Flash serves as \texttt{CompilerLM}. The training data contain 3,368 SFT trajectories and 608 GRPO training
% instances. We train
% \texttt{RouterLM} using Low-Rank Adaptation (LoRA)~\citep{hu2021lora}. Qwen3.5-9B is trained and evaluated with its
% thinking mode disabled.
Semantic retrieval uses BGE-M3, lexical retrieval uses BM25, and relational
operations execute read-only SQLite queries. For \ours, Gemini~3.5 Flash serves
as \texttt{CompilerLM}. Training uses 3,368 SFT trajectories and 608 GRPO
instances. We train \texttt{RouterLM} with LoRA~\citep{hu2021lora}.
Qwen3.5-9B has thinking mode disabled.

% All methods are evaluated on the same test questions and associated source
% contexts. Gemini~3.5 Flash serves as the shared \texttt{AnswerLM} for all methods. All outputs are scored by the same frozen
% Gemini~3.5 Flash judge using an identical evaluation prompt. Language-model calls use temperature 0, and multi-round methods are limited to six rounds and 4,000 characters of retained evidence. 
% We repeat each in-domain experiment
% three times and report mean task success rate, with standard deviations
% provided in Appendix~\ref{sec:appendix-main-std}. Additional implementation, prompt, training, and evaluation details are provided in
% Appendices~\ref{sec:appendix_ours_training} and ~\ref{sec:appendix_prompts}.

All methods are evaluated on same test questions and source contexts.
Gemini~3.5 Flash serves as the shared \texttt{AnswerLM} for all methods. All
outputs are scored by the same frozen Gemini~3.5 Flash judge using an identical
prompt. LLM calls use temperature 0, and multi-round methods are
limited to six rounds and 4,000 characters of retained evidence. We repeat each
in-domain experiment three times and report mean task success rate. Standard deviations, prompt, additional implementation, training, and evaluation details are provided in Appendices~\ref{sec:appendix-main-std}, ~\ref{sec:appendix_ours_training},
and~\ref{sec:appendix_prompts}.

\subsection{\ours Performance}
\label{sec:performance}

Table~\ref{tab:main-results} compares success rates of \ours and baselines across six benchmarks.
Appendices~\ref{sec:appendix_operation_analysis}, \ref{sec:appendix_cost}, and~\ref{sec:appendix_failure} provide operation-usage, cost, and failure analyses. There are four main takeaways:

\begin{table*}[t]
\centering
\small
\caption{\textbf{In-domain success rates (\%) of \ours across six benchmark families.} The
listed LM directly generates the program for Direct Code, guides evidence
retrieval for IRCoT and Interact-RAG, and serves as \texttt{RouterLM} for
\ours. Bold denotes the best result in each column.}
%\vspace{2pt}
\label{tab:main-results}
\setlength{\tabcolsep}{1.8pt}

\begin{tabular}{@{}l>{\hspace{5pt}}l@{\hspace{0pt}}cccccc|@{\hspace{5pt}}c@{\hspace{2.5pt}}}
\toprule
Method & LM
& DataBench & FinQA & HiTab & HotpotQA & LaMP
& \shortstack{MultiHiertt} & \textbf{Avg.} \\
\midrule

\rowcolor{gray!20}
\multicolumn{9}{@{}l}{\textbf{\textit{Retrieval baselines}}} \\
Direct
& --
& $6.7$ & $1.0$ & $23.0$ & $41.7$ & $69.7$ & $1.0$ & $23.8$ \\
Fixed Retrieval
& --
& $36.0$ & $27.3$ & $72.3$ & $79.3$ & $64.0$ & $55.3$ & $55.7$ \\

\midrule
\rowcolor{gray!20}
\multicolumn{9}{@{}l}{\textbf{\textit{Direct program generation}}} \\
Direct Code
& Qwen3.5-9B
& $80.7$ & $1.0$ & $26.0$ & $64.0$ & $54.7$ & $3.3$ & $38.3$  \\
Direct Code
& Gemini~3.5 Flash
& $88.0$ & $5.0$ & $46.0$ & $78.3$ & $48.0$ & $5.0$ & $45.1$ \\

\midrule
\rowcolor{gray!20}
\multicolumn{9}{@{}l}{\textbf{\textit{Iterative and agentic retrieval}}} \\
IRCoT
& Qwen3.5-9B
& $36.0$ & $27.0$ & $64.0$ & $94.0$ & $65.0$ & $56.0$ & $57.0$ \\
IRCoT
& Gemini~3.5 Flash
& $39.3$ & $30.0$ & $63.7$ & $\mathbf{94.7}$ & $67.3$ & $58.0$ & $58.8$ \\
Interact-RAG
& Qwen3.5-9B
& $30.0$ & $45.0$ & $43.0$ & $87.0$ & $63.0$ & $56.0$ & $54.0$ \\
Interact-RAG
& Gemini~3.5 Flash
& $36.7$ & $52.7$ & $72.7$ & $79.7$ & $62.3$ & $54.0$ & $59.7$ \\

\midrule
\rowcolor{gray!20}
\multicolumn{9}{@{}l}{\textbf{\textit{\ours}}} \\
\ours
& Qwen3.5-9B
& $86.0$ & $57.0$ & $60.7$ & $83.0$ & $68.3$ & $29.3$ & $64.1$ \\
\ours
& Gemini~3.5 Flash
& $\mathbf{90.0}$ & $58.3$ & $67.3$ & $86.0$ & $\mathbf{71.7}$ & $50.3$ & $70.6$ \\
\ours
& Qwen3.5-9B (SFT+GRPO)
& $83.3$ & $\mathbf{68.3}$ & $\mathbf{76.3}$ & $92.0$
& $68.3$ & $\mathbf{65.3}$ & $\mathbf{75.6}$ \\

\bottomrule
\end{tabular}
%\vspace{-15pt}
\end{table*}

\textbf{First, \ours achieves the strongest overall performance.} The
SFT+GRPO-trained system obtains an average success rate of $75.6\%$, a
$15.9\%$ improvement over the strongest  large-LLM-
controlled  baseline, Gemini-based Interact-RAG at
$59.7\%$. This result reflects strong performance across diverse source
formats and evidence requirements. Re-evaluating predictions with two
alternative LLM judges and token-level F1 preserves the overall performance
advantage (Appendix~\ref{sec:appendix_eval_robustness}).

\textbf{Second, the design of \ours provides substantial gains even without \texttt{RouterLM} training.} With Qwen3.5-9B, training-free \ours achieves $64.1\%$,
compared with $57.0\%$ for the strongest Qwen-based baseline. With Gemini~3.5
Flash, training-free \ours reaches $70.6\%$, compared with $59.7\%$ for the
strongest Gemini-based baseline. Notably, training-free \ours with Qwen3.5-9B as \texttt{RouterLM} also
outperforms IRCoT and Interact-RAG using Gemini~3.5 Flash as their retrieval
controller. By supporting adaptive retrieval, computation, and evidence assessment, \ours
provides gains beyond those obtained by equipping the retrieval baselines with
a stronger controller.

\textbf{Third, training substantially strengthens the evidence-routing
capability of the smaller model.} SFT and GRPO improve Qwen3.5-9B from
$64.1\%$ to $75.6\%$. The small, trained Qwen router also
outperforms the large, training-free Gemini~3.5 Flash router by $5.0\%$ and
achieves higher performance on four of the six benchmarks. Thus, SFT and GRPO enable a smaller,
trainable model to surpass a larger black-box model by learning specialized
evidence-routing behavior.

\textbf{Fourth, heterogeneous tasks benefit from adaptive evidence construction
rather than a single strategy.} The strongest program generation and
agentic retrieval baselines exhibit different strengths across benchmarks. Gemini-based Direct Code excels on DataBench but obtains only $45.1\%$ on average, whereas Gemini-based Interact-RAG obtains $59.7\%$ with a clear advantage on FinQA, HiTab, and MultiHiertt. In comparison, \ours reaches
$75.6\%$ and maintains more balanced performance across the six benchmark
families. By adaptively supporting primitive retrieval and computation together
with synthesized code, \ours provides a more flexible evidence-construction
process across heterogeneous tasks.

\subsection{Ablation Studies}
\label{sec:ablations}

We ablate the evidence-operation space, the GRPO reward and data-selection
strategy, and the contributions of the two training stages.
Table~\ref{tab:framework-ablation} summarizes the results. There are three main takeaways:

\begin{table*}[t]
\centering
\caption{\textbf{Success rates (\%) of \ours ablations across six
benchmark families.} The variants evaluate the evidence-operation space,
training stages, GRPO reward, and GRPO data selection. Bold denotes the best result in each column.}
%\vspace{2pt}
\label{tab:framework-ablation}
\small
\setlength{\tabcolsep}{3.5pt}

\begin{tabularx}{\textwidth}{
    @{}
    >{\raggedright\arraybackslash}X
    rrrrrr
    |r
    @{}
}
\toprule
Variant & DataBench & FinQA & HiTab & HotpotQA & LaMP & MultiHiertt & \textbf{Avg.} \\
\midrule

\rowcolor{gray!20}
\multicolumn{8}{@{}l}{\textbf{\textit{Evidence-operation space}}} \\

\ours$_{\texttt{Train-Free, Primitive Only}}$
& $25.3$ & $61.0$ & $63.7$ & $80.7$ & $67.7$ & $42.7$ & $56.8$ \\

\ours$_{\texttt{Train-Free, Synthesis Only}}$
& $85.3$ & $45.0$ & $54.3$ & $88.7$ & $63.0$ & $23.0$ & $59.9$ \\

\ours$_{\texttt{Train-Free, Full}}$
& $\mathbf{86.0}$ & $57.0$ & $60.7$ & $83.0$ & $68.3$ & $29.3$ & $64.1$ \\

\midrule
\rowcolor{gray!20}
\multicolumn{8}{@{}l}{\textbf{\textit{GRPO reward and data selection}}} \\

\ours$_{\texttt{SFT+GRPO, Accuracy Only}}$
& $82.0$ & $58.0$ & $65.0$ & $87.3$ & $66.0$ & $55.0$ & $68.9$ \\

\ours$_{\texttt{SFT+GRPO, Unfiltered}}$
& $83.7$ & $62.0$ & $74.0$ & $85.3$ & $\mathbf{69.3}$ & $63.7$ & $73.0$ \\

\midrule
\rowcolor{gray!20}
\multicolumn{8}{@{}l}{\textbf{\textit{Training stages}}} \\

\ours$_{\texttt{SFT}}$
& $82.0$ & $55.7$ & $68.3$ & $82.7$ & $66.3$ & $49.0$ & $67.3$ \\

\ours$_{\texttt{GRPO}}$
& $81.7$ & $64.0$ & $75.7$ & $88.3$ & $64.0$ & $61.7$ & $72.6$ \\

\ours$_{\texttt{SFT+GRPO}}$
& $83.3$ & $\mathbf{68.3}$ & $\mathbf{76.3}$
& $\mathbf{92.0}$ & $68.3$ & $\mathbf{65.3}$ & $\mathbf{75.6}$ \\

\bottomrule
\end{tabularx}
%\vspace{-17pt}
\end{table*}

\textbf{First, primitive and synthesized operations provide complementary
evidence-construction capabilities.} Primitive-only \ours achieves $56.8\%$,
and synthesis-only $59.9\%$. Neither operation type is uniformly superior across families,
indicating that effectiveness depends on the task's evidence
requirements. Combining both yields the strongest training-free average of $64.1\%$,
outperforming either variant and demonstrating the benefit of a
broader evidence-operation space.

\textbf{Second, reward shaping and selective GRPO data construction
contribute to performance.} Using answer correctness as the only reward reduces the average from $75.6\%$
to $68.9\%$, showing that partial answer quality and structural validity provide useful distinctions among trajectories. Removing
the mixed-outcome group filter yields $73.0\%$, indicating that groups
containing both successful and unsuccessful trajectories provide a stronger
relative learning signal. Because filtering produces a smaller training
pool, its higher performance demonstrates improved sample efficiency.

\textbf{Third, SFT and GRPO provide complementary improvements.} Relative to
training-free \ours at $64.1\%$, SFT increases the average to $67.3\%$, while
GRPO increases it to $72.6\%$. Combining both yields the strongest result at $75.6\%$, exceeding
either stage and demonstrating their complementary benefits. SFT establishes valid operation selection and formulation from
filtered successful trajectories, while GRPO optimizes
routing trajectories using downstream outcomes.

\subsection{Generalization}
\label{sec:generalization}

\subsubsection{Generalization to unseen benchmarks}

Table~\ref{tab:unseen-benchmarks} reports zero-shot performance on three
held-out benchmarks: 2WikiMultiHopQA~\citep{ho2020constructing},
TAT-QA~\citep{zhu2021tat}, and
WikiTableQuestions~\citep{pasupat2015compositional}. These benchmarks cover
multi-hop text question answering, hybrid text-table numerical reasoning, and
table question answering, respectively. Here, Direct Code, IRCoT, and Interact-RAG use the pretrained Qwen3.5-9B as their program-generation or retrieval-guidance model, while \ours uses the SFT+GRPO-trained Qwen3.5-9B \texttt{RouterLM}. There are two main takeaways.
\textbf{First, \ours generalizes consistently across all three held-out
benchmarks.} It achieves an average success rate of $79.3\%$, outperforming the
strongest baseline at $64.3\%$ by $15.0\%$. The improvements over
the strongest baseline on each benchmark are $6.0\%$ on 2WikiMultiHopQA,
$8.0\%$ on TAT-QA, and $15.0\%$ on WikiTableQuestions. \textbf{Second, the learned routing policy generalizes across heterogeneous
source representations.} The improvements span collections of Wikipedia
passages, financial reports combining text and tables, and semi-structured
Wikipedia tables. These results show that evidence-routing behavior learned
from the in-domain benchmarks transfers to new task distributions and source
organizations.

\subsubsection{Transfer across \texttt{CompilerLM}s}

We keep the trained \texttt{RouterLM} fixed and replace the default
Gemini~3.5 Flash \texttt{CompilerLM} with Gemini~3.5 Flash Lite and
Qwen3.5-9B, neither of which is used as a compiler during training.
Table~\ref{tab:compiler-transfer} reports the results. There are two main takeaways. \textbf{First, the learned routing policy
transfers effectively to a lighter \texttt{CompilerLM}.} Gemini~3.5 Flash Lite
achieves $73.2\%$ in-domain, a decrease of only $2.4\%$ from the default
compiler, while matching its held-out performance at $79.3\%$. \textbf{Second, the routing policy also transfers across model families and
substantially different model scales.} With Qwen3.5-9B as
\texttt{CompilerLM}, \ours achieves $67.2\%$ in-domain and $73.3\%$ on the
held-out benchmarks, remaining $7.5\%$ and $9.0\%$ above the strongest
evaluated baselines, respectively. These results show that the learned
\texttt{RouterLM} does not depend on a single compiler model and can retain
strong performance with substantially more compact compilers.

\begin{table}[ht]
\centering
\small
\setlength{\tabcolsep}{4pt}
%\vspace{-13pt}
\caption{\textbf{Zero-shot success rate (\%) of \ours across three unseen benchmarks.}}
%\vspace{2pt}
\label{tab:unseen-benchmarks}
\begin{tabular}{lrrr|r}
\toprule
Method & 2Wiki & TAT-QA & WTQ & Avg. \\
\midrule
Direct & $72.0$ & $1.0$ & $10.0$ & $27.7$ \\
Direct Code & $45.0$ & $38.0$ & $59.0$ & $47.3$  \\
Fixed Retrieval & $58.0$ & $66.0$ & $52.0$ & $58.7$ \\
IRCoT & $69.0$ & $73.0$ & $51.0$ & $64.3$ \\
Interact-RAG & $77.0$ & $57.0$ & $48.0$ & $60.7$ \\
\ours$_{\texttt{SFT+GRPO}}$ & $\mathbf{83.0}$ & $\mathbf{81.0}$ & $\mathbf{74.0}$ & $\mathbf{79.3}$ \\
\bottomrule
\end{tabular}
%\vspace{-15pt}
\end{table}

\begin{table}[ht]
\centering
\small
%\vspace{-4pt}
\caption{\textbf{\ours in-domain and held-out success rates (\%) with alternative
\texttt{CompilerLM}s.}}
%\vspace{2pt}
\label{tab:compiler-transfer}
\begin{tabular}{lrr}
\toprule
\texttt{CompilerLM} & In-domain avg. & Held-out avg. \\
\midrule
Qwen3.5-9B & $67.2$ & $73.3$ \\
Gemini 3.5 Flash Lite & $73.2$ & $\mathbf{79.3}$ \\
Gemini 3.5 Flash & $\mathbf{75.6}$ & $\mathbf{79.3}$ \\
\bottomrule
\end{tabular}
%\vspace{-15pt}
\end{table}

\section{Conclusion}
\label{sec:conclusion}

We introduced \ours, a learned framework that broadens context construction from retrieving source content to adaptively deriving task-relevant evidence through retrieval and computation. Trained through SFT and GRPO, a lightweight \texttt{RouterLM} iteratively formulates primitive operations or requests customized code, assesses their outcomes, and determines when the evidence is sufficient for final inference. Across six in-domain benchmarks, \ours outperforms retrieval-based, agentic, and direct program-generation approaches on average. It generalizes to three held-out benchmarks and retains strong performance with smaller code-generation models. Our ablations show that primitive and synthesized operations provide complementary capabilities and that learning strengthens the evidence-construction policy. These results demonstrate the effectiveness and generalizability of adaptive evidence construction across heterogeneous tasks and sources.

\paragraph{Limitations.}
Constructing the training data for \ours relies on execution-derived outcomes, which introduces additional computational overhead during data collection. At inference time, iterative calls to \texttt{RouterLM} and occasional code synthesis may also increase latency and computational cost relative to single-pass retrieval. Future work could explore more efficient trajectory construction and cost-aware routing policies that jointly optimize task performance and computational efficiency.

\subsection*{AI use statement}

We used generative AI tools in three ways. First, as writing
aids: we provided draft text for grammatical correction and structural
refinement, and the authors verified and edited all outputs. Second, for
retrieval and discovery, such as finding potentially related work. The authors read all cited papers and verified every reference. Third, to assist research execution, including environment setup and experiment code debugging. All AI-assisted code was reviewed and tested by the authors, and we take full responsibility for the content of this paper.

\subsection*{Ethics statement}

This work improves how language models construct evidence from heterogeneous sources. All experiments use publicly available benchmarks, and no new human-subject data was collected. Because \ours executes model-generated code, relational queries are read-only and synthesized programs run in an isolated environment. When applied to private or sensitive data, deployments should enforce similar sandboxing along with appropriate access controls and privacy safeguards.

\subsection*{Reproducibility statement}

For reproducibility, we describe the full \ours framework and training
procedure in Section~\ref{sec:method}. Dataset construction, SFT and GRPO
hyperparameters, the GRPO objective, and the reward definition are detailed in
Appendix~\ref{sec:appendix_ours_training}, and standard deviations over three
runs are reported in Appendix~\ref{sec:appendix-main-std}. Baseline and variant
implementations are described in Appendices~\ref{sec:appendix-baselines}
and~\ref{sec:appendix-ablations}, and all prompts are provided in
Appendix~\ref{sec:appendix_prompts}. Our code, data, and trained \texttt{RouterLM} checkpoint will be publicly
released under an open-source license upon acceptance.

\bibliography{iclr2027_conference}
%\bibliographystyle{iclr2027_conference}

%\clearpage
\newpage
\appendix

\startcontents[appendices]

\section*{Appendix--RECAST: Learning to Compute the Right Context through Adaptive Evidence Routing}

\printcontents[appendices]{}{1}{}

\newpage
\section{Benchmarks}
\subsection{Benchmark Descriptions}
\label{sec:appendix_bench_desc}
We use six benchmark families for training and in-domain evaluation and three
additional benchmarks for zero-shot evaluation.

\paragraph{Training and in-domain benchmarks:}\mbox{}

$\bullet$ \textbf{DataBench.}
DataBench evaluates question answering over real-world dataframes from diverse
application domains~\citep{grijalba2024question}. Questions require analyzing
the provided table and returning either a scalar or a list. 
% We use DataBench Lite, where each question is paired with a representative 20-row sample of the corresponding table.

$\bullet$ \textbf{FinQA.}
FinQA contains numerical reasoning questions grounded in financial reports
composed of prose passages and structured tables~\citep{chen2021finqa}.
Answering a question may require identifying relevant information from either
source and performing arithmetic over the extracted values.

$\bullet$ \textbf{HiTab.}
HiTab evaluates question answering over cross-domain tables with hierarchical
row and column headers~\citep{cheng2022hitab}. Its questions cover lookup,
comparison, aggregation, and numerical reasoning while requiring the table's
multi-level structure to be correctly interpreted.

$\bullet$ \textbf{HotpotQA.}
HotpotQA is a multi-hop question-answering benchmark built from Wikipedia
articles~\citep{yang2018hotpotqa}. 
% We use its distractor setting, where each question is accompanied by multiple passages and the required evidence may span more than one article.

$\bullet$ \textbf{LaMP.}
LaMP evaluates personalized language modeling using a current input and a
profile containing historical items from the same user~\citep{salemi2024lamp}.
We combine LaMP-1, 2, 3, 4, 5, and 7 into one benchmark family, covering
citation identification, movie tagging, product rating, news headline
generation, scholarly title generation, and tweet paraphrasing. We exclude
LaMP-6 because it relies on the restricted Avocado email corpus.

$\bullet$ \textbf{MultiHiertt.}
MultiHiertt contains numerical reasoning questions grounded in financial
reports with multiple hierarchical tables and accompanying text
passages~\citep{zhao2022multihiertt}. Questions may require combining evidence
across multiple tables or between tabular and textual content.

\paragraph{Held-out benchmarks:}\mbox{}

$\bullet$ \textbf{2WikiMultiHopQA.}
2WikiMultiHopQA evaluates multi-hop question answering over collections of
Wikipedia passages~\citep{ho2020constructing}. Its questions require reasoning
across multiple documents through compositional, comparison, or inference
relations.

$\bullet$ \textbf{TAT-QA.}
TAT-QA contains questions grounded in financial reports that combine prose
passages and tables~\citep{zhu2021tat}. It includes span-based, counting, and
arithmetic questions, with answers expressed using different numerical scales.

$\bullet$ \textbf{WikiTableQuestions.}
WikiTableQuestions evaluates compositional question answering over
semi-structured Wikipedia tables~\citep{pasupat2015compositional}. Questions
may require selecting, comparing, or aggregating values from multiple table
entries.

\newpage
\subsection{Benchmark Examples}
\label{sec:appendix_bench_ex}

The following examples illustrate the task and source format of each benchmark.
Because complete sources may contain many records, we show only shortened,
representative snippets. The benchmarks span heterogeneous sources, including
text passages, flat and hierarchical tables, financial reports, dataframes,
and user profiles. Their questions also require different forms of evidence,
ranging from direct retrieval and multi-hop linking to aggregation, numerical
reasoning, and personalized prediction or generation.

\begin{benchmarkexample}{DataBench}
(*@\textbf{Query:}@*)
Which month sees the highest number of complaints?

(*@\textbf{Source snippet (CSV):}@*)
complaint_type,borough,hour,month_name,agency
HEAT/HOT WATER,MANHATTAN,18,March,HPD
...
[additional dataframe rows omitted]

(*@\textbf{Answer:}@*)
January
\end{benchmarkexample}

\begin{benchmarkexample}{FinQA}
(*@\textbf{Query:}@*)
What percentage of the contractual obligations due in 2008 are
maturities of long-term debt?

(*@\textbf{Source snippet (financial report):}@*)
Contractual obligations for future payments at December 31, 2007
were as follows (in millions):

                                      2008    2009    2010
Maturities of long-term debt           267   1,300   1,069
Debt obligations with right of offset 2,013   2,013   2,013
...

(*@\textbf{Answer:}@*)
11%
\end{benchmarkexample}

\begin{benchmarkexample}{HiTab}
(*@\textbf{Query:}@*)
In 2018, which club did Russell sign with Major League Soccer?

(*@\textbf{Source snippet (hierarchical table row):}@*)
{
  "row_headers": ["sporting kansas city"],
  "values": [
    {"column_headers": ["season"], "value": "2018"},
    {"column_headers": ["league", "division"],
     "value": "major league soccer"},
    ...
  ]
}

(*@\textbf{Answer:}@*)
["sporting kansas city"]
\end{benchmarkexample}

\begin{benchmarkexample}{HotpotQA}
(*@\textbf{Query:}@*)
Where is the head office of the airway on whose Board of Directors
Wanjiku Mugane serves?

(*@\textbf{Source snippet (Wikipedia passages):}@*)
Title: Wanjiku Mugane
"... member of the board of directors of Kenya Airways ..."

Title: Kenya Airways
"The carrier's head office is located in Embakasi, Nairobi,
with its hub at Jomo Kenyatta International Airport."

(*@\textbf{Answer:}@*)
Embakasi, Nairobi
\end{benchmarkexample}

\begin{benchmarkexample}{LaMP-1}
(*@\textbf{Query:}@*)
For the author of "RoboCare: Pervasive Intelligence for the
Domestic Care of the Elderly," which reference is related?

[1] Crowd analysis: a survey
[2] Differential-drive in-pipe robot for moving inside
    urban gas pipelines

(*@\textbf{Source snippet (user profile):}@*)
{
  "title": "Temporary maps for robust localization in
            semi-static environments",
  "abstract": "Accurate and robust localization is essential
               for autonomous mobile robots. ..."
}

(*@\textbf{Answer:}@*)
[2]
\end{benchmarkexample}

\begin{benchmarkexample}{MultiHiertt}
(*@\textbf{Query:}@*)
What is the average amount of Balance forward of SquareFeet
and 2015 of Mexico UnitCount?

(*@\textbf{Source snippet (hierarchical HTML table):}@*)
<tr>
  <td>Balance forward</td>
  <td></td>
  <td></td>
  <td>632</td>
  <td>84,382</td>
</tr>
...
[additional table required by the question omitted]

(*@\textbf{Answer:}@*)
43336
\end{benchmarkexample}

\begin{benchmarkexample}{2WikiMultiHopQA}
(*@\textbf{Query:}@*)
Does Lizzie Jelfs have the same nationality as Paul S. Devrouax?

(*@\textbf{Source snippet (Wikipedia passages):}@*)
Title: Lizzie Jelfs
"Elizabeth Coulson (nee Jelfs) is a British former
professional tennis player."

Title: Paul S. Devrouax
"Paul S. Devrouax was an American architect."

(*@\textbf{Answer:}@*)
no
\end{benchmarkexample}

\begin{benchmarkexample}{TAT-QA}
(*@\textbf{Query:}@*)
What is the percentage change in cash and short-term marketable
securities from 2018 to 2019?

(*@\textbf{Source snippet (financial table row):}@*)
{
  "item": "Cash and short-term marketable securities",
  "2019": "$454",
  "2018": "$524",
  "2017": "$656"
}

(*@\textbf{Answer:}@*)
-13.36%
\end{benchmarkexample}

\begin{benchmarkexample}{WikiTableQuestions}
(*@\textbf{Query:}@*)
Which actor has the most NAACP Image Awards?

(*@\textbf{Source snippet (Wikipedia table):}@*)
Year | Supporting Actor | Motion Picture
1995 | Al Freeman, Jr.   | Malcolm X
1996 | Laurence Fishburne| Higher Learning
1997 | Samuel L. Jackson | A Time to Kill
1998 | Morgan Freeman    | Amistad
...
[additional table rows omitted]

(*@\textbf{Answer:}@*)
Morgan Freeman
\end{benchmarkexample}

\newpage
\section{\ours}
\label{sec:appendix_ours} 

\subsection{\ours Training Details}
\label{sec:appendix_ours_training} 
\paragraph{Dataset.}
We begin with 4,660 unique questions from the six training benchmark families.
For each question, we collect two initial candidate trajectories. A third
trajectory is generated only when both initial attempts fail, and a fourth is
generated only when all preceding attempts fail. This adaptive collection
procedure produces 11,563 candidate trajectories in total. Trajectory
collection uses a pretrained Qwen3.5-9B \texttt{RouterLM}, BGE-M3 for semantic
retrieval, and Gemini~3.5 Flash as \texttt{CompilerLM}, \texttt{AnswerLM}, and
the correctness judge.

We retain only trajectories that produce a correct final answer, complete
without parsing or execution failure, satisfy the action schema, and conform to
the final \ours action space. When several valid candidates are available for
the same question, we retain the trajectory with stronger native or exact
answer agreement, followed by higher token F1 and lower execution and token
cost. This process
produces 3,064 questions with expert trajectory. We reserve approximately 10\% of these questions (306 questions) for validation, leaving 2,758 unique questions for SFT.
We apply capped upsampling by repeating examples from underrepresented
benchmark families and empirical difficulty levels, with at most two copies of
any example. The resulting SFT corpus contains 3,368 complete trajectories.

For GRPO, we begin with 1,200 questions that are neither consistently solved
nor entirely unsolved during initial trajectory collection. Each has at least one
valid expert and at least one failed attempt. From the selected SFT checkpoint,
we sample four new trajectories per question and apply
\emph{mixed-outcome filtering}, retaining only groups containing at least one
successful and one unsuccessful trajectory. After structural validation and
benchmark-family balancing through repetition, the GRPO training set contains
608 prompt rows derived from 567 unique questions. A separate set of 300
questions, with 50 from each benchmark family, is used for GRPO validation.
Table~\ref{tab:training-data-composition} summarizes the resulting datasets.

\begin{table}[th]
\centering
\small
\caption{\textbf{Composition of the trajectory collection and the final SFT
and GRPO training sets.}}
\label{tab:training-data-composition}
\setlength{\tabcolsep}{4pt}
\begin{tabular}{lrrrr}
\toprule
Family & Questions & Raw traj. & SFT rows & GRPO rows \\
\midrule
DataBench  & 885 & 2,080 & 665 & 97  \\
FinQA      & 410 & 1,109 & 349 & 101 \\
HiTab      & 935 & 2,381 & 567 & 96  \\
HotpotQA   & 785 & 1,674 & 660 & 96  \\
LaMP       & 735 & 1,814 & 519 & 95  \\
MultiHiertt & 910 & 2,505 & 608 & 123 \\
\midrule
Total & 4,660 & 11,563 & 3,368 & 608 \\
\bottomrule
\end{tabular}
\end{table}

\paragraph{SFT for \ours$_{\texttt{SFT+GRPO}}$.}
We fine-tune Qwen3.5-9B using LoRA~\citep{hu2021lora} on eight A100 80GB
GPUs. LoRA is applied to all linear layers with rank 32, scaling parameter
$\alpha_{\mathrm{LoRA}}=64$, and dropout 0.05. We train for two epochs
with a global batch size of 16, a maximum sequence length of 16,384 tokens, and
a learning rate of $5\times10^{-5}$. The loss supervises all
\texttt{RouterLM} decisions across each multi-round trajectory, and we select
the checkpoint with the best development performance.

\paragraph{GRPO for \ours$_{\texttt{SFT+GRPO}}$.}
We continue LoRA fine-tuning from the selected SFT checkpoint using eight A100
80GB GPUs. GRPO uses a learning rate of $2\times10^{-6}$, a batch size of
eight prompts, and four sampled routing trajectories per prompt. Trajectories
are sampled with temperature 0.8 and top-$p=0.95$. We train the model for three
epochs.

\paragraph{GRPO Objective.}
For each prompt $x$, GRPO~\citep{shao2024deepseekmath} samples a group of
$G=4$ trajectories $\{y_i\}_{i=1}^{G}$ with rewards
$\{r_i\}_{i=1}^{G}$. The group-relative advantage is
\begin{equation}
    A_i =
    \frac{r_i-\bar{r}}
    {\operatorname{std}(r_1,\ldots,r_G)+\delta},
    \qquad
    \bar{r}=\frac{1}{G}\sum_{j=1}^{G}r_j,
\end{equation}
where $\delta$ is a small constant for numerical stability. Let
\begin{equation}
    \rho_{i,t}(\theta)=
    \frac{\pi_{\theta}(y_{i,t}\mid x,y_{i,<t})}
    {\pi_{\mathrm{old}}(y_{i,t}\mid x,y_{i,<t})}
\end{equation}
denote the importance ratio at token $t$. The optimization objective is
\begin{align}
\mathcal{J}_{\mathrm{GRPO}}(\theta)
={}&
\mathbb{E}_{i,t}\!\left[
\min\!\left(
\rho_{i,t}(\theta)A_i,\,
\operatorname{clip}
\!\left(\rho_{i,t}(\theta),1-\epsilon,1+\epsilon\right)A_i
\right)\right]
\nonumber\\
&-\beta
D_{\mathrm{KL}}^{\mathrm{low\mbox{-}var}}
\!\left(\pi_{\theta}\,\|\,\pi_{\mathrm{ref}}\right),
\end{align}
where $\pi_{\mathrm{old}}$ is the rollout policy and
$\pi_{\mathrm{ref}}$ is the fixed reference policy. We use
$\epsilon=0.2$ and $\beta=0.01$. The terminal trajectory reward is applied to
all generated tokens.

% \paragraph{GRPO Reward Explanation.}
% The terminal reward for a routing trajectory $\tau$ is
% \begin{equation}
%     r(\tau)
%     =
%     0.90\,C(\tau)
%     +0.08\,F_1(\tau)
%     +0.02\,V(\tau),
% \end{equation}
% where $C(\tau)\in\{0,1\}$ is final-answer correctness determined by the frozen
% judge, $F_1(\tau)\in[0,1]$ is the token-level F1 between the normalized
% generated and reference answers, and $V(\tau)\in\{0,1\}$ indicates that the
% trajectory follows the required action schema and completes without a pipeline
% failure. The weights are chosen so that final-answer correctness dominates the reward:
% every correct trajectory receives a higher reward than any incorrect
% trajectory. Token-level F1 and structural validity serve only as bounded
% secondary signals for distinguishing trajectories with the same correctness
% outcome. The accuracy-only ablation uses $r(\tau)=C(\tau)$.

\paragraph{GRPO Reward Explanation.}
The terminal reward for a routing trajectory $\tau$ is:
\begin{equation}
    R(\tau)
    =
    0.90\,C(\tau)
    +0.08\,F(\tau)
    +0.02\,V(\tau),
\end{equation}
where $C(\tau)\in\{0,1\}$ is final-answer correctness determined by the frozen
judge, $F(\tau)\in[0,1]$ is the token-level F1 between the normalized
generated and reference answers, and $V(\tau)\in\{0,1\}$ indicates that the
trajectory follows the required action schema and completes without a pipeline
failure. The weights are chosen so that final-answer correctness dominates the
reward: every correct trajectory receives a higher reward than any incorrect
trajectory. Token-level F1 and structural validity serve only as bounded
secondary signals for distinguishing trajectories with the same correctness
outcome. Although these secondary terms carry only $10\%$ of the reward weight, they
are the only signal that separates trajectories with the same correctness
outcome. With $R(\tau)=C(\tau)$, all correct trajectories in a group receive
identical advantages, as do all incorrect ones, so GRPO cannot distinguish a
near-miss answer from a malformed or failed trajectory. The accuracy-only ablation uses $R(\tau)=C(\tau)$.

\newpage

\subsection{\ours Main-Result Performance with Standard Deviations}
\label{sec:appendix-main-std}

Table~\ref{tab:main-results-std} reports the sample standard deviations corresponding
to the mean success rates in Table~\ref{tab:main-results}. Statistics are
computed over three complete evaluations of the same 600-question test set.
Although all language-model calls use temperature~0, outputs can still vary
across runs due to nondeterminism in hosted model APIs and batched GPU
inference. In multi-round methods, such variation can change an intermediate
routing or retrieval decision and propagate to later rounds. Across methods,
the overall standard deviation ranges from 0.3\% to 0.8\%, indicating stable
aggregate performance across repeated evaluations.

% Table~\ref{tab:main-results-std} reports the sample standard deviations
% corresponding to the mean success rates in
% Table~\ref{tab:main-results}. Statistics are computed over three complete
% evaluations of the same 600-question test set. Across methods, the overall standard deviation ranges from $0.3\%$ to $0.8\%$,
% indicating stable aggregate performance across repeated evaluations.

\begin{table*}[ht]
\centering
\small
\caption{\textbf{Standard deviations of in-domain success rates (\%) across
three evaluations.} The corresponding mean results are reported in
Table~\ref{tab:main-results}.}
%\vspace{2pt}
\label{tab:main-results-std}
\setlength{\tabcolsep}{1.8pt}

\begin{tabular}{@{}l>{\hspace{5pt}}l@{\hspace{0pt}}cccccc|@{\hspace{5pt}}c@{\hspace{2.5pt}}}
\toprule
Method & LM
& DataBench & FinQA & HiTab & HotpotQA & LaMP
& \shortstack{MultiHiertt} & \textbf{Avg.} \\
\midrule

\rowcolor{gray!20}
\multicolumn{9}{@{}l}{\textbf{\textit{Retrieval baselines}}} \\
Direct
& --
& $1.2$ & $0.0$ & $1.0$ & $3.8$ & $1.5$ & $0.0$ & $0.4$ \\
Fixed Retrieval
& --
& $1.0$ & $1.2$ & $0.6$ & $1.5$ & $1.7$ & $1.5$ & $0.3$ \\

\midrule
\rowcolor{gray!20}
\multicolumn{9}{@{}l}{\textbf{\textit{Direct program generation}}} \\
Direct Code
& Qwen3.5-9B
& $0.6$ & $1.0$ & $1.0$ & $0.0$ & $1.5$ & $0.6$ & $0.5$ \\
Direct Code
& Gemini~3.5 Flash
& $1.7$ & $2.0$ & $1.7$ & $0.6$ & $1.0$ & $0.0$ & $0.4$ \\

\midrule
\rowcolor{gray!20}
\multicolumn{9}{@{}l}{\textbf{\textit{Iterative and agentic retrieval}}} \\
IRCoT
& Qwen3.5-9B
& $2.0$ & $1.5$ & $0.6$ & $0.0$ & $1.0$ & $1.2$ & $0.8$ \\
IRCoT
& Gemini~3.5 Flash
& $0.6$ & $1.0$ & $1.2$ & $0.6$ & $1.5$ & $3.5$ & $0.8$ \\
Interact-RAG
& Qwen3.5-9B
& $1.5$ & $2.3$ & $2.5$ & $0.6$ & $1.5$ & $1.5$ & $0.7$ \\
Interact-RAG
& Gemini~3.5 Flash
& $2.9$ & $2.1$ & $2.3$ & $2.1$ & $1.5$ & $1.7$ & $0.3$ \\

\midrule
\rowcolor{gray!20}
\multicolumn{9}{@{}l}{\textbf{\textit{\ours}}} \\
\ours
& Qwen3.5-9B
& $1.0$ & $1.0$ & $2.5$ & $3.6$ & $0.6$ & $1.5$ & $0.5$ \\
\ours
& Gemini~3.5 Flash
& $1.0$ & $1.5$ & $2.3$ & $1.0$ & $2.5$ & $1.5$ & $0.5$ \\
\ours
& Qwen3.5-9B (SFT+GRPO)
& $1.5$ & $3.1$ & $0.6$ & $2.6$ & $2.3$ & $2.5$ & $0.5$ \\

\bottomrule
\end{tabular}
%\vspace{-10pt}
\end{table*}

\newpage
\subsection{Evidence-Construction Behavior}
\label{sec:appendix_operation_analysis}

Table~\ref{tab:operation-usage} summarizes the trained \ours policy's
operation usage and routing rounds on the
600-question in-domain test set. We report primitive and synthesis calls
per question, the percentage of questions invoking synthesis, and average
routing rounds. Routing rounds include evidence-operation requests and
context-acceptance decisions.

\begin{table}[ht]
\centering
\small
\caption{\textbf{Operation usage and routing rounds of trained \ours.}
Statistics are computed from one evaluation run, with 100 questions per
benchmark. Calls and rounds are averaged per question; synthesis usage
denotes the percentage of questions with at least one synthesis call.}
\label{tab:operation-usage}
\setlength{\tabcolsep}{5pt}
\begin{tabular}{lrrrr}
\toprule
Benchmark
& \shortstack{Primitive\\calls}
& \shortstack{Synthesis\\calls}
& \shortstack{Synthesis\\usage (\%)}
& \shortstack{Routing\\rounds} \\
\midrule
DataBench   & 0.04 & 1.12 & 98.0 & 2.16 \\
FinQA       & 3.00 & 0.22 & 21.0 & 4.22 \\
HiTab       & 2.60 & 0.07 &  6.0 & 3.67 \\
HotpotQA    & 1.46 & 0.00 &  0.0 & 2.43 \\
LaMP        & 2.89 & 0.04 &  4.0 & 3.93 \\
MultiHiertt & 2.11 & 0.40 & 35.0 & 3.50 \\
\midrule
Overall     & 2.02 & 0.31 & 27.3 & 3.32 \\
\bottomrule
\end{tabular}
\end{table}

The learned policy exhibits distinct operation-use patterns across
benchmarks. DataBench predominantly uses synthesized computation,
whereas HotpotQA relies entirely on primitives in this run. FinQA and
MultiHiertt combine substantial primitive use with synthesis on a subset
of questions, while HiTab and LaMP invoke synthesis less frequently.
Overall, synthesis is used on 27.3\% of questions, showing that the
framework does not universally delegate evidence construction to
\texttt{CompilerLM}. Routing length also varies across benchmarks,
from 2.16 rounds on DataBench to 4.22 on FinQA. These observations
complement the operation-space ablations by showing how the learned
policy uses different evidence operations and interaction lengths
across heterogeneous tasks.

\newpage
\subsection{Robustness to Evaluation Choices}
\label{sec:appendix_eval_robustness}

Because Gemini~3.5 Flash provides correctness judgments during both
training and evaluation, we examine whether the reported improvements
persist under alternative evaluators. We re-score the saved predictions
from one inference replicate per method on the same 600-question
in-domain test set, without regenerating answers or retraining models.
We use GPT-4.1 (\texttt{gpt-4.1-2025-04-14}) and DeepSeek-V4-Flash
(\texttt{deepseek-ai/DeepSeek-V4-Flash-0731}) with the same correctness
prompt as the original judge. We additionally compute token-level F1
using SQuAD-style normalization.

\paragraph{Alternative LLM judges.}
Tables~\ref{tab:main-results-gpt41}
and~\ref{tab:main-results-deepseek} report the re-evaluated success rates.
Trained \ours achieves $74.8\%$ under GPT-4.1 and $72.8\%$ under
DeepSeek-V4-Flash, exceeding the strongest baseline by $15.8\%$ and
$15.3\%$, respectively. Although absolute scores vary across judges,
\ours remains the strongest method overall under both evaluators.
These results indicate that its performance advantage is not specific
to evaluation by the model used for training-time correctness judgments.

\begin{table*}[ht]
\centering
\small
\caption{\textbf{In-domain success rates (\%) judged by GPT-4.1.} Same runs as Table~\ref{tab:main-results} (one replicate per row), re-judged with \texttt{gpt-4.1-2025-04-14} using the identical judge prompt. Bold denotes the best result in each column.}
% \vspace{2pt}
\label{tab:main-results-gpt41}
\setlength{\tabcolsep}{1.8pt}
\begin{tabular}{@{}l>{\hspace{5pt}}l@{\hspace{0pt}}cccccc|@{\hspace{5pt}}c@{\hspace{2.5pt}}}
\toprule
Method & LM
& DataBench & FinQA & HiTab & HotpotQA & LaMP
& \shortstack{MultiHiertt} & \textbf{Avg.} \\
\midrule
\rowcolor{gray!20}
\multicolumn{9}{@{}l}{\textbf{\textit{Retrieval baselines}}} \\
Direct
& --
& $7.0$ & $1.0$ & $21.0$ & $39.0$ & $\mathbf{73.0}$ & $1.0$ & $23.7$ \\
Fixed Retrieval
& --
& $35.0$ & $26.0$ & $68.0$ & $79.0$ & $68.0$ & $57.0$ & $55.5$ \\
\midrule
\rowcolor{gray!20}
\multicolumn{9}{@{}l}{\textbf{\textit{Direct program generation}}} \\
Direct Code
& Qwen3.5-9B
& $70.0$ & $3.0$ & $21.0$ & $62.0$ & $57.0$ & $0.0$ & $35.5$ \\
Direct Code
& Gemini~3.5 Flash
& $77.0$ & $7.0$ & $41.0$ & $78.0$ & $47.0$ & $5.0$ & $42.5$ \\
\midrule
\rowcolor{gray!20}
\multicolumn{9}{@{}l}{\textbf{\textit{Iterative and agentic retrieval}}} \\
IRCoT
& Qwen3.5-9B
& $33.0$ & $27.0$ & $61.0$ & $94.0$ & $68.0$ & $57.0$ & $56.7$ \\
IRCoT
& Gemini~3.5 Flash
& $36.0$ & $28.0$ & $59.0$ & $\mathbf{96.0}$ & $70.0$ & $53.0$ & $57.0$ \\
Interact-RAG
& Qwen3.5-9B
& $30.0$ & $44.0$ & $39.0$ & $87.0$ & $65.0$ & $55.0$ & $53.3$ \\
Interact-RAG
& Gemini~3.5 Flash
& $38.0$ & $53.0$ & $70.0$ & $78.0$ & $62.0$ & $53.0$ & $59.0$ \\
\midrule
\rowcolor{gray!20}
\multicolumn{9}{@{}l}{\textbf{\textit{\ours}}} \\
\ours
& Qwen3.5-9B
& $85.0$ & $54.0$ & $58.0$ & $77.0$ & $69.0$ & $29.0$ & $62.0$ \\
\ours
& Gemini~3.5 Flash
& $\mathbf{86.0}$ & $59.0$ & $62.0$ & $85.0$ & $69.0$ & $51.0$ & $68.7$ \\
\ours
& Qwen3.5-9B (SFT+GRPO)
& $84.0$ & $\mathbf{69.0}$ & $\mathbf{73.0}$ & $88.0$ & $67.0$ & $\mathbf{68.0}$ & $\mathbf{74.8}$ \\
\bottomrule
\end{tabular}
% \vspace{-15pt}
\end{table*}

\begin{table*}[ht]
\centering
\small
\caption{\textbf{In-domain success rates (\%) judged by DeepSeek-V4-Flash.} Same runs as Table~\ref{tab:main-results} (one replicate per row), re-judged with \texttt{deepseek-ai/DeepSeek-V4-Flash-0731} using the identical judge prompt and decoding at temperature 0. Bold denotes the best result in each column.}
% \vspace{2pt}
\label{tab:main-results-deepseek}
\setlength{\tabcolsep}{1.8pt}
\begin{tabular}{@{}l>{\hspace{5pt}}l@{\hspace{0pt}}cccccc|@{\hspace{5pt}}c@{\hspace{2.5pt}}}
\toprule
Method & LM
& DataBench & FinQA & HiTab & HotpotQA & LaMP
& \shortstack{MultiHiertt} & \textbf{Avg.} \\
\midrule
\rowcolor{gray!20}
\multicolumn{9}{@{}l}{\textbf{\textit{Retrieval baselines}}} \\
Direct
& --
& $7.0$ & $1.0$ & $19.0$ & $38.0$ & $68.0$ & $1.0$ & $22.3$ \\
Fixed Retrieval
& --
& $36.0$ & $26.0$ & $66.0$ & $77.0$ & $62.0$ & $56.0$ & $53.8$ \\
\midrule
\rowcolor{gray!20}
\multicolumn{9}{@{}l}{\textbf{\textit{Direct program generation}}} \\
Direct Code
& Qwen3.5-9B
& $79.0$ & $2.0$ & $23.0$ & $64.0$ & $56.0$ & $2.0$ & $37.7$ \\
Direct Code
& Gemini~3.5 Flash
& $\mathbf{87.0}$ & $6.0$ & $42.0$ & $78.0$ & $45.0$ & $5.0$ & $43.8$ \\
\midrule
\rowcolor{gray!20}
\multicolumn{9}{@{}l}{\textbf{\textit{Iterative and agentic retrieval}}} \\
IRCoT
& Qwen3.5-9B
& $35.0$ & $25.0$ & $59.0$ & $92.0$ & $63.0$ & $55.0$ & $54.8$ \\
IRCoT
& Gemini~3.5 Flash
& $37.0$ & $26.0$ & $56.0$ & $\mathbf{94.0}$ & $67.0$ & $53.0$ & $55.5$ \\
Interact-RAG
& Qwen3.5-9B
& $30.0$ & $42.0$ & $40.0$ & $87.0$ & $60.0$ & $55.0$ & $52.3$ \\
Interact-RAG
& Gemini~3.5 Flash
& $38.0$ & $48.0$ & $68.0$ & $77.0$ & $61.0$ & $53.0$ & $57.5$ \\
\midrule
\rowcolor{gray!20}
\multicolumn{9}{@{}l}{\textbf{\textit{\ours}}} \\
\ours
& Qwen3.5-9B
& $86.0$ & $51.0$ & $56.0$ & $78.0$ & $65.0$ & $29.0$ & $60.8$ \\
\ours
& Gemini~3.5 Flash
& $\mathbf{87.0}$ & $57.0$ & $60.0$ & $84.0$ & $\mathbf{69.0}$ & $50.0$ & $67.8$ \\
\ours
& Qwen3.5-9B (SFT+GRPO)
& $83.0$ & $\mathbf{63.0}$ & $\mathbf{70.0}$ & $88.0$ & $66.0$ & $\mathbf{67.0}$ & $\mathbf{72.8}$ \\
\bottomrule
\end{tabular}
% \vspace{-15pt}
\end{table*}

\paragraph{Agreement between judges.}
Tables~\ref{tab:judge-agreement-gpt41}
and~\ref{tab:judge-agreement-deepseek} compare the alternative judges
with Gemini on identical predictions. We report raw agreement and Cohen's $\kappa$, which adjusts agreement
for the level expected by chance from each judge's correct/incorrect
labeling rates. Across methods, Gemini--GPT-4.1
agreement ranges from $95.2\%$ to $98.7\%$, with Cohen's $\kappa$ between $0.90$ and $0.96$. Gemini--DeepSeek agreement ranges from $95.3\%$ to $98.8\%$, with $\kappa$ between $0.88$ and $0.98$.
For trained \ours, agreement is $96.7\%$ with GPT-4.1 and $95.3\%$
with DeepSeek. Both alternative judges assign equal or lower success rates than Gemini, but preserve the overall performance advantage.
The Gemini scores here correspond to the selected replicate rather
than the three-run means in Table~\ref{tab:main-results}.

\begin{table}[ht]
\centering
\small
\caption{\textbf{Agreement between the Gemini judge and GPT-4.1.} Average success rate (\%) under each judge on the same 600 predictions per row, raw agreement, and Cohen's $\kappa$.}
\label{tab:judge-agreement-gpt41}
\setlength{\tabcolsep}{3pt}
\begin{tabular}{@{}llcccc@{}}
\toprule
Method & LM & Gemini & GPT-4.1 & Agree (\%) & $\kappa$ \\
\midrule
Direct & -- & $23.7$ & $23.7$ & $98.7$ & $0.96$ \\
Fixed Retrieval & -- & $56.0$ & $55.5$ & $96.8$ & $0.94$ \\
Direct Code & Qwen3.5-9B & $37.7$ & $35.5$ & $95.2$ & $0.90$ \\
Direct Code & Gemini~3.5 Flash & $44.7$ & $42.5$ & $96.5$ & $0.93$ \\
IRCoT & Qwen3.5-9B & $57.0$ & $56.7$ & $97.0$ & $0.94$ \\
IRCoT & Gemini~3.5 Flash & $58.2$ & $57.0$ & $95.8$ & $0.91$ \\
Interact-RAG & Qwen3.5-9B & $54.0$ & $53.3$ & $96.7$ & $0.93$ \\
Interact-RAG & Gemini~3.5 Flash & $59.5$ & $59.0$ & $96.8$ & $0.93$ \\
\ours & Qwen3.5-9B & $63.5$ & $62.0$ & $97.2$ & $0.94$ \\
\ours & Gemini~3.5 Flash & $70.5$ & $68.7$ & $96.2$ & $0.91$ \\
\ours & Qwen3.5-9B (SFT+GRPO) & $76.2$ & $74.8$ & $96.7$ & $0.91$ \\
\bottomrule
\end{tabular}
\end{table}

\begin{table}[ht]
\centering
\small
\caption{\textbf{Agreement between the Gemini judge and DeepSeek-V4-Flash.} Average success rate (\%) under each judge on the same 600 predictions per row, raw agreement, and Cohen's $\kappa$.}
\label{tab:judge-agreement-deepseek}
\setlength{\tabcolsep}{3pt}
\begin{tabular}{@{}llcccc@{}}
\toprule
Method & LM & Gemini & DeepSeek-V4-Flash & Agree (\%) & $\kappa$ \\
\midrule
Direct & -- & $23.7$ & $22.3$ & $98.0$ & $0.94$ \\
Fixed Retrieval & -- & $56.0$ & $53.8$ & $96.5$ & $0.93$ \\
Direct Code & Qwen3.5-9B & $37.7$ & $37.7$ & $97.0$ & $0.94$ \\
Direct Code & Gemini~3.5 Flash & $44.7$ & $43.8$ & $98.8$ & $0.98$ \\
IRCoT & Qwen3.5-9B & $57.0$ & $54.8$ & $97.2$ & $0.94$ \\
IRCoT & Gemini~3.5 Flash & $58.2$ & $55.5$ & $95.3$ & $0.90$ \\
Interact-RAG & Qwen3.5-9B & $54.0$ & $52.3$ & $97.3$ & $0.95$ \\
Interact-RAG & Gemini~3.5 Flash & $59.5$ & $57.5$ & $96.7$ & $0.93$ \\
\ours & Qwen3.5-9B & $63.5$ & $60.8$ & $97.0$ & $0.94$ \\
\ours & Gemini~3.5 Flash & $70.5$ & $67.8$ & $96.0$ & $0.91$ \\
\ours & Qwen3.5-9B (SFT+GRPO) & $76.2$ & $72.8$ & $95.3$ & $0.88$ \\
\bottomrule
\end{tabular}
\end{table}

\newpage
\paragraph{Token-level F1.}
Table~\ref{tab:main-results-f1} provides a complementary evaluation
without an LLM judge. Trained \ours obtains an average token-level F1
of $57.3\%$, compared with $45.3\%$ for the strongest baseline.
Although token-level F1 contributes to the training reward, its weight
is only $0.08$, compared with $0.90$ for final-answer correctness.
The improvement therefore extends beyond the primary training objective.
Because token-level F1 measures lexical overlap and can penalize
semantically equivalent answers, we report it as a supplementary metric
rather than a replacement for correctness evaluation.

\begin{table*}[ht]
\centering
\small
\caption{\textbf{In-domain token-level F1 (\%) across six benchmark families.} Computed on the same runs as Table~\ref{tab:main-results-gpt41} with SQuAD-style normalization. Bold denotes the best result in each column.}
% \vspace{2pt}
\label{tab:main-results-f1}
\setlength{\tabcolsep}{1.8pt}
\begin{tabular}{@{}l>{\hspace{5pt}}l@{\hspace{0pt}}cccccc|@{\hspace{5pt}}c@{\hspace{2.5pt}}}
\toprule
Method & LM
& DataBench & FinQA & HiTab & HotpotQA & LaMP
& \shortstack{MultiHiertt} & \textbf{Avg.} \\
\midrule
\rowcolor{gray!20}
\multicolumn{9}{@{}l}{\textbf{\textit{Retrieval baselines}}} \\
Direct
& --
& $6.7$ & $0.7$ & $8.7$ & $37.5$ & $53.9$ & $0.0$ & $17.9$ \\
Fixed Retrieval
& --
& $38.6$ & $11.8$ & $47.1$ & $72.6$ & $56.8$ & $32.9$ & $43.3$ \\
\midrule
\rowcolor{gray!20}
\multicolumn{9}{@{}l}{\textbf{\textit{Direct program generation}}} \\
Direct Code
& Qwen3.5-9B
& $69.5$ & $0.0$ & $20.3$ & $61.2$ & $40.0$ & $0.0$ & $31.8$ \\
Direct Code
& Gemini~3.5 Flash
& $60.9$ & $1.7$ & $31.2$ & $73.3$ & $43.0$ & $1.7$ & $35.3$ \\
\midrule
\rowcolor{gray!20}
\multicolumn{9}{@{}l}{\textbf{\textit{Iterative and agentic retrieval}}} \\
IRCoT
& Qwen3.5-9B
& $41.6$ & $11.8$ & $40.0$ & $86.3$ & $57.5$ & $32.5$ & $45.0$ \\
IRCoT
& Gemini~3.5 Flash
& $39.8$ & $13.7$ & $39.7$ & $\mathbf{88.4}$ & $\mathbf{58.2}$ & $31.8$ & $45.3$ \\
Interact-RAG
& Qwen3.5-9B
& $31.7$ & $22.4$ & $27.6$ & $80.8$ & $50.0$ & $30.8$ & $40.6$ \\
Interact-RAG
& Gemini~3.5 Flash
& $40.2$ & $23.6$ & $47.3$ & $71.1$ & $50.8$ & $28.5$ & $43.6$ \\
\midrule
\rowcolor{gray!20}
\multicolumn{9}{@{}l}{\textbf{\textit{\ours}}} \\
\ours
& Qwen3.5-9B
& $78.0$ & $27.2$ & $38.9$ & $73.0$ & $52.6$ & $12.9$ & $47.1$ \\
\ours
& Gemini~3.5 Flash
& $\mathbf{82.8}$ & $31.6$ & $41.1$ & $80.8$ & $\mathbf{58.2}$ & $29.5$ & $54.0$ \\
\ours
& Qwen3.5-9B (SFT+GRPO)
& $81.2$ & $\mathbf{33.6}$ & $\mathbf{51.2}$ & $82.3$ & $54.2$ & $\mathbf{41.2}$ & $\mathbf{57.3}$ \\
\bottomrule
\end{tabular}
% \vspace{-15pt}
\end{table*}
% \newpage

\newpage
\section{Baselines and \ours Variants}

All baselines and \ours variants follow the shared experimental settings
described in the main text. We provide their method-specific implementation
details below.

\subsection{Baselines}
\label{sec:appendix-baselines}

\paragraph{Direct.}
\texttt{AnswerLM} receives only the task and produces the final solution
without accessing its associated source. This baseline measures performance
without context construction.

\paragraph{Fixed Retrieval.}
The task is used directly as the query for a single BGE-M3 retrieval call.
The three highest-ranked text chunks are selected and passed to
\texttt{AnswerLM}, subject to the shared evidence limit. The method performs
neither query reformulation nor additional retrieval rounds.

\paragraph{Direct Code.}
The designated LLM generates a Python program for the task in a single call.
The program is executed over the complete source, and its output is passed to
the shared \texttt{AnswerLM}. Unlike \ours, Direct Code does not select between
evidence operations, assess intermediate evidence, or revise its computation
over multiple rounds. We evaluate both Qwen3.5-9B and Gemini~3.5 Flash as the
program-generating LLM.

\paragraph{IRCoT.}
IRCoT~\citep{trivedi2023interleaving} alternates retrieval with
sentence-level reasoning. At each round, it retrieves three BGE-M3 chunks using
either the original task or the latest generated factual sentence as the
query. Newly retrieved evidence is added to the working context before the
next reasoning step. We retain at most 15 evidence items, use the original
IRCoT stopping procedure, and allow at most six rounds. The accumulated
evidence is passed to the shared \texttt{AnswerLM} for final solution
generation.

\paragraph{Interact-RAG.}
Interact-RAG~\citep{hui2026interact} uses a Global Planner, Adaptive Reasoner,
and Executor to iteratively construct and refine retrieval requests. The
Executor may issue up to two searches per round, combining BGE-M3 semantic
retrieval with optional BM25 matching and constraints over entities or source
records. Each search returns between three and five results, and the workflow
runs for at most six rounds before passing its evidence to the shared
\texttt{AnswerLM}. 

We adapt IRCoT and Interact-RAG to the per-instance sources,
models, and retrieval components used in our evaluation while preserving their
central reasoning and control procedures.

\subsection{\ours Variants}
\label{sec:appendix-ablations}

\paragraph{Training-free \ours.}
We evaluate the complete \ours framework with either a pretrained
Qwen3.5-9B or Gemini~3.5 Flash \texttt{RouterLM}. Both variants can invoke the
lexical, semantic, and relational primitives, request synthesized code, and
assess the resulting evidence over multiple rounds. Neither
\texttt{RouterLM} receives task-specific training. All remaining framework
components are identical to those used by the trained model.

\paragraph{Operation-space ablations.}
\ours$_{\texttt{Primitive Only}}$ removes the \texttt{Synthesize} action and
allows the training-free Qwen3.5-9B \texttt{RouterLM} to use only lexical,
semantic, and relational primitives. Conversely,
\ours$_{\texttt{Synthesis Only}}$ removes \texttt{Call\_Primitive}, requiring
the same router to construct evidence exclusively through synthesized programs.
These variants isolate the contributions of the two evidence-operation
families.

% \paragraph{Training-stage ablations.}
% \ours$_{\texttt{SFT}}$ uses the selected SFT checkpoint without subsequent
% policy optimization. \ours$_{\texttt{GRPO}}$ initializes directly from the
% pretrained Qwen3.5-9B model and applies GRPO using the same mixed-outcome
% training pool and reward as the full model.
% \ours$_{\texttt{SFT+GRPO}}$ initializes GRPO from the selected SFT checkpoint
% and constitutes the complete trained framework. The primitive operations,
% \texttt{CompilerLM}, \texttt{AnswerLM}, execution environment, and inference
% budgets remain fixed across these variants.

\paragraph{SFT for \ours$_{\texttt{SFT}}$.}
We fine-tune pretrained Qwen3.5-9B for five epochs using the same training
data and hyperparameters as the SFT stage of
\ours$_{\texttt{SFT+GRPO}}$. We select the checkpoint with the best
development performance across the five epochs.

\paragraph{GRPO for \ours$_{\texttt{GRPO}}$.}
We apply GRPO directly to pretrained Qwen3.5-9B without SFT initialization.
Training uses the same mixed-outcome question pool, shaped reward, LoRA
configuration, and optimization settings as the GRPO stage of
\ours$_{\texttt{SFT+GRPO}}$. We use eight A100 80GB GPUs and train for
three epochs.

\paragraph{GRPO reward and data ablations.}
\ours$_{\texttt{Accuracy Only}}$ retains the SFT initialization,
mixed-outcome training pool, and GRPO configuration but replaces the shaped
reward with binary final-answer correctness. 
\ours$_{\texttt{Unfiltered}}$ retains the SFT initialization and shaped reward
but trains on all 1,200 GRPO candidate questions without mixed-outcome
filtering. 
% The number of optimization updates is held fixed to isolate the effect of GRPO data selection.

\paragraph{\texttt{CompilerLM} variants.}
The default \ours configuration uses Gemini~3.5 Flash as
\texttt{CompilerLM}. To evaluate transfer across code-generation models, we
replace it with Gemini~3.5 Flash Lite or Qwen3.5-9B while keeping the trained
\texttt{RouterLM}, \texttt{AnswerLM}, primitive operations, and execution
environment unchanged. The alternative compiler is invoked only when
\texttt{RouterLM} selects \texttt{Synthesize}.

\newpage
\section{Cost Analysis}
\label{sec:appendix_cost}

Table~\ref{tab:baseline-token-cost} presents average task success rates
alongside mean inference token usage per question. Token counts include
input and output tokens across routing, code generation, and final
answering, but exclude the correctness judge used for evaluation.
We separate Qwen3.5-9B and Gemini~3.5 Flash tokens to distinguish
computation performed by the smaller and larger models.

\begin{table*}[ht]
\centering
\small
\caption{\textbf{Average task success rates (\%) and inference token usage
per question, in thousands (k).} Token counts include input and output
tokens across all inference calls and exclude the evaluation judge.
The LM column follows Table~\ref{tab:main-results}.
Totals are computed before rounding.}
%\vspace{2pt}
\label{tab:baseline-token-cost}
\setlength{\tabcolsep}{4pt}
\begin{tabular}{@{}llrrrr@{}}
\toprule
Method & LM & Success (\%) & Qwen (k) & Gemini (k) & Total (k) \\
\midrule

\rowcolor{gray!20}
\multicolumn{6}{@{}l}{\textbf{\textit{Retrieval baselines}}} \\
Direct & --
& $23.8$ & $0.0$ & $0.6$ & $0.6$ \\
Fixed Retrieval & --
& $55.7$ & $0.0$ & $1.6$ & $1.6$ \\

\midrule
\rowcolor{gray!20}
\multicolumn{6}{@{}l}{\textbf{\textit{Direct program generation}}} \\
Direct Code & Qwen3.5-9B
& $38.3$ & $4.7$ & $0.0$ & $4.7$ \\
Direct Code & Gemini~3.5 Flash
& $45.1$ & $0.0$ & $4.8$ & $4.8$  \\

\midrule
\rowcolor{gray!20}
\multicolumn{6}{@{}l}{\textbf{\textit{Iterative and agentic retrieval}}} \\
IRCoT & Qwen3.5-9B
& $57.0$ & $5.1$ & $1.7$ & $6.8$ \\
IRCoT & Gemini~3.5 Flash
& $58.8$ & $0.0$ & $10.0$ & $10.0$ \\
Interact-RAG & Qwen3.5-9B
& $54.0$ & $14.6$ & $1.8$ & $16.5$ \\
Interact-RAG & Gemini~3.5 Flash
& $59.7$ & $0.0$ & $26.9$ & $26.9$ \\

\midrule
\rowcolor{gray!20}
\multicolumn{6}{@{}l}{\textbf{\textit{\ours}}} \\
\ours & Qwen3.5-9B
& $64.1$ & $10.6$ & $7.1$ & $17.7$ \\
\ours & Gemini~3.5 Flash
& $70.6$ & $0.0$ & $31.8$ & $31.8$ \\
\ours & Qwen3.5-9B (SFT+GRPO)
& $\mathbf{75.6}$ & $13.9$ & $5.5$ & $19.4$ \\

\bottomrule
\end{tabular}
\end{table*}

Trained \ours achieves higher success with lower token usage than
Gemini-based Interact-RAG, obtaining 75.6\% versus 59.7\% success while
using 27.9\% fewer total tokens and 79.5\% fewer Gemini tokens.
Compared with training-free \ours using a Gemini \texttt{RouterLM},
it also improves success from 70.6\% to 75.6\% while reducing total
token usage from 31.8k to 19.4k. These results illustrate the benefit
of assigning iterative evidence construction to a trained smaller model,
with larger models supporting compilation and final answering.

% There are two main observations.
% \textbf{First, trained \ours achieves higher success with lower token
% usage than Gemini-based agentic retrieval.} \ours obtains 75.6\% success,
% compared with 59.7\% for Gemini-based Interact-RAG, while using
% 27.9\% fewer total tokens and 79.5\% fewer Gemini tokens.
% Compared with training-free \ours using a Gemini \texttt{RouterLM},
% the trained Qwen model also improves success from 70.6\% to 75.6\%
% while reducing total token usage from 31.8k to 19.4k.
% These results support assigning iterative evidence construction to a
% trained smaller model while using the larger model for compilation
% and final answering.

% \textbf{Second, additional inference computation can support stronger
% task performance, but greater token usage alone is insufficient.}
% Fixed Retrieval uses only 1.6k tokens per question and achieves 55.7\%
% success, providing an inexpensive reference point. Direct Code and
% IRCoT also use fewer tokens than trained \ours, but achieve lower
% average success rates. Within \ours, training improves success from
% 64.1\% to 75.6\% with an increase in total token usage from 17.7k to
% 19.4k, while Gemini usage decreases from 7.1k to 5.5k. The resulting
% trade-off reflects both the amount of computation and how it is
% allocated across models and evidence operations.

\newpage
\section{Failure Case Analysis}
\label{sec:appendix_failure}

We qualitatively inspect unsuccessful trajectories to identify recurring
difficulties and limitations in evidence construction. Direct serves as a
reference without source access; the analysis below focuses on methods
that retrieve or compute information from the supplied sources.

\subsection{Baselines}

\paragraph{Fixed Retrieval.}
Fixed Retrieval selects evidence through a single similarity-based search.
The selected content is passed directly to the answer model without
subsequent assessment or revision. This limits its ability to address
information needs that become apparent only after inspecting the initial
results, such as identifying a missing connection between sources or
obtaining additional context to interpret a table entry.

\paragraph{IRCoT.}
In the inspected Qwen-based trajectories, intermediate reasoning frequently
repeats earlier statements, while subsequent retrieval returns little new
information. An initial interpretation can therefore persist across
rounds without being adequately reconsidered. Additional reasoning and
retrieval steps do not necessarily translate into progress toward resolving
the remaining information need.

\paragraph{Interact-RAG.}
The inspected Qwen-based trajectories exhibit both early termination and
repeated search refinement with limited progress. Early termination and
exhaustion of the round budget are more common among unsuccessful
trajectories, although neither alone establishes the cause of an error.
These patterns highlight the difficulty of determining whether the evidence
is sufficient and whether further searches are likely to be informative.

\paragraph{Direct Code.}
Given only the source profile and a few sampled records, many generated programs, particularly from Gemini~3.5 Flash, inspect the source instead of computing an answer: they print records without assigning the required output, so no result is returned. Other failures include invalid Python syntax, mostly from Qwen3.5-9B programs truncated after extended reasoning written as code comments, and runtime errors such as accessing missing record fields.
These cases show that direct program generation must satisfy both the
task's computational requirements and the execution interface.
The single-call setup does not allow the model to revise an unsuccessful
program in response to execution feedback.

\subsection{\ours}

A recurring failure pattern is obtaining relevant evidence that does not
fully resolve the question. The router may locate the correct topic or an
intermediate entity without obtaining the required relation or constraint.
Other cases involve interpreting the requested computation, such as choosing
an absolute rather than relative difference, even when the synthesized
program executes correctly. These observations suggest that more precise
evidence requests and sufficiency assessment are promising directions for
further improving the learned policy.

\newpage
\section{Prompts}
\label{sec:appendix_prompts}

We present the system prompts and input formats for \texttt{RouterLM},
\texttt{CompilerLM}, \texttt{AnswerLM}, and the correctness judge.

% Requires the listings package.
\lstdefinestyle{recastprompt}{
    basicstyle=\ttfamily\footnotesize,
    breaklines=true,
    breakatwhitespace=true,
    columns=fullflexible,
    keepspaces=true,
    showstringspaces=false,
    frame=single,
    framerule=0.3pt,
    framesep=5pt,
    aboveskip=8pt,
    belowskip=8pt
}

\subsection{\texttt{RouterLM}}
\label{sec:appendix_router_prompt}

The system prompt defines the action space, primitive interfaces, and
requirements for formulating evidence operations.

\begin{lstlisting}[style=recastprompt]
You are a low-cost adaptive evidence router. Prepare compact, source-grounded context for a separate large executor. Do not answer the task.

Choose exactly one action per round and return JSON only.

Call one primitive:
{
  "action": "CALL_PRIMITIVE",
  "evidence_goal": "evidence or computed result this call must produce",
  "primitive": "relational|lexical|semantic",
  "arguments": {},
  "reason": "brief reason"
}

Request one-shot code compilation:
{
  "action": "SYNTHESIZE",
  "level": "code",
  "evidence_goal": "the exact deterministic evidence operation",
  "required_output": "computed or selected evidence with source IDs",
  "reason": "why a single primitive is insufficient"
}

Accept context:
{
  "action": "ACCEPT_CONTEXT",
  "reason": "why the retained evidence already satisfies the task"
}

Primitive engines:
- relational: Run one read-only SQLite SELECT over normalized records. Use it for exact filtering, joins, grouping, aggregation, sorting, arithmetic, and comparison. Use only columns in the metadata profile. Return record_id or source_record_ids; aggregates can use GROUP_CONCAT(record_id) AS source_record_ids.
  Arguments: {"query": "SELECT ...", "scope_record_ids": [... optional ...]}.
- lexical: BM25 keyword retrieval over raw record text. Use it for exact names, identifiers, dates, error codes, quotations, and distinctive vocabulary.
  Arguments: {"query": "...", "top_k": N, "scope_record_ids": [... optional ...]}.
- semantic: Dense-vector retrieval over record text. Use it for paraphrases, concepts, abstract similarity, and high-level relevance. It requires an embedding model and reports a failure if none is configured.
  Arguments: {"query": "...", "top_k": N, "scope_record_ids": [... optional ...]}.

All engines return evidence with source IDs, counts, truncation information, and diagnostics. A scope restricts an engine to records selected earlier.

Only call primitives listed in metadata_profile.available_primitives.

Use relational directly when one SQL query can perform exact filtering, arithmetic, grouping, aggregation, sorting, comparison, or joins over the normalized source. It queries only metadata_profile.sql.table and its listed columns; it does not open or query a database, spreadsheet, or other raw file merely because that file is represented by a source record. Use code to query or join data stored inside attached files.

Use code when the requested operation requires parsing or an algorithm that cannot be naturally expressed by one available primitive, such as schema normalization, a sliding window, graph traversal, regex validation, or processing an irregular raw file.

If the task asks for an exact computed result, retrieved operands alone are not sufficient. Request the calculation before accepting. Conversely, do not request synthesis merely to summarize retrieved prose or predict a label. For demonstration or personalization tasks, retrieve representative or contrastive examples and let the executor perform the final judgment.

The compiler receives the complete source and accumulated evidence, but sees only your synthesis specification rather than the original task. Therefore make evidence_goal and required_output complete and unambiguous. Repeat every constraint needed for exact implementation, including units, grouping keys, ordering, comparison direction, and boundary inclusivity or exclusivity. Inspect execution feedback and retry only with a materially different operation or implementation level.
\end{lstlisting}

\subsection{\texttt{CompilerLM}}
\label{sec:appendix_compiler_prompt}

\texttt{CompilerLM} receives the router's synthesis specification rather
than the original task. Its input includes the source profile, source
samples, accumulated evidence, and recent operation outcomes. The generated
program can access the complete source through the execution environment.

\begin{lstlisting}[style=recastprompt]
Mechanically implement the supplied router specification as one temporary Python evidence program. You do not receive the original user task. Do not add new reasoning goals or produce a natural-language answer.

Return JSON only:
{
  "name": "short_program_name",
  "description": "evidence produced",
  "code": "complete Python statements that assign the final value to result"
}

The code is unrestricted:
- Any valid Python statements and installed packages are allowed.
- You may define functions, parsers, algorithms, or data structures.
- You may call any available primitive through tools.
- Raw record contents can be opened through metadata.runtime_path.

Runtime variables:
- task: a JSON serialization of the router's synthesis specification, not the original user task.
- source_records: every source record. Each has top-level record_id, text, and metadata. For file-backed records, text is empty at execution time; open metadata.runtime_path for complete contents. For inline records, when text is a JSON object, its parsed fields are nested under content.
- accumulated_evidence: retained evidence wrappers. Each wrapper has origin, content, and source_record_ids. The retrieved source row is nested under the wrapper's content key.
- profile: the deterministic metadata profile.
- tools.call(name, **arguments), tools.relational(...), tools.lexical(...), and tools.semantic(...).

The installed-library description is in metadata_profile.code_runtime. Imports are not restricted, but a package must actually be installed in that runtime. Prefer its listed libraries or the Python standard library.

Use exact paths shown in metadata_profile and bounded_source_samples. Never move a nested field to the top level by assumption: for example, a profile field named content.label is read from row["content"]["label"], while the same retrieved row is read from evidence_item["content"]["content"]["label"]. Raw CSV, TSV, and text records remain plain strings: do not call json.loads on them. Open metadata.runtime_path with csv, pandas, or an appropriate parser based on metadata.suffix. Only use json.loads when the content is valid JSON. metadata.absolute_path identifies the host source but is not readable inside the isolated container. To locate a named file, match metadata.relative_path, then open that same record's metadata.runtime_path. Never attempt to open metadata.absolute_path. Implement every stated constraint literally, including units, grouping keys, sort direction, and strict versus inclusive comparisons. Do not silently weaken or reinterpret the router specification.

Bounded evidence and samples can be truncated. Process complete source_records or runtime_path files whenever completeness matters.

The request can include previous_code_attempts selected by the router after a failed execution. Each contains the exact earlier program and its runtime feedback. Repair the smallest relevant defect while preserving any correct parsing, computation, and full-source scope. In particular, a provenance-only failure should be repaired by copying source_records[i]["record_id"] into the result, not by replacing complete-source processing with a bounded evidence preview.

Assign compact evidence to result. If required_output specifies field names, labels, ordering, or an artifact schema, preserve them literally in the evidence. Every output must include non-empty source_record_ids copied from the records actually used; a separate variable named source_record_ids is not enough. Do not hard-code an expected answer or invent provenance.
\end{lstlisting}

\subsection{\texttt{AnswerLM}}
\label{sec:appendix_answer_prompt}

\texttt{AnswerLM} receives the original task and the constructed evidence.
The source contract describes the source representation and its fields.

\begin{lstlisting}[style=recastprompt]
Answer the task using the accepted evidence and its source contract. The  evidence may contain facts, computed values, or demonstrations whose patterns must be applied to the task. Perform the final reasoning yourself.

Return only the concise final answer without explanation. If the evidence is genuinely insufficient or contradictory, return exactly: I don't know
\end{lstlisting}

The user message has the following structure.

\begin{lstlisting}[style=recastprompt]
{
  "task": "<original task>",
  "source_contract": {
    "description": "<source description>",
    "representation": "<source representation>",
    "fields": "<source fields>"
  },
  "accepted_evidence": ["<constructed evidence items>"]
}
\end{lstlisting}

\subsection{Correctness Judge}
\label{sec:appendix_judge_prompt}

The same correctness-judging prompt is used across methods. The judge
compares the candidate answer with the reference answer; it does not
participate in evidence construction.

\begin{lstlisting}[style=recastprompt]
You are a strict answer-correctness evaluator. Determine whether a candidate answer correctly answers the question when compared with the reference answer. Accept semantically equivalent aliases, harmless formatting differences, and equivalent dates or numbers. Reject answers that contradict the reference, identify the wrong entity, omit a required part, or add a material false claim.

Return exactly one JSON object and no surrounding prose: {"correct": true, "reason": "brief explanation"}
\end{lstlisting}

The user message contains the question, reference answer, and candidate
answer.

\begin{lstlisting}[style=recastprompt]
{
  "question": "<original question>",
  "reference_answer": "<reference answer>",
  "candidate_answer": "<model prediction>"
}
\end{lstlisting}

\end{document}